%% file: vakra_main.tex
\documentclass[11pt]{article}

\usepackage[preprint]{acl}
\usepackage{todonotes}
\usepackage{booktabs}
\usepackage{times}
\usepackage{amsmath}
\usepackage{latexsym}
\usepackage{makecell}
\usepackage{multirow}
\usepackage{multicol}
\usepackage[T1]{fontenc}
\usepackage[utf8]{inputenc}

\usepackage{microtype}

\usepackage{inconsolata}
\usepackage{graphicx}
\usepackage{listings}
\usepackage{xcolor}

\lstdefinestyle{pycode}{
  language=Python,
  basicstyle=\ttfamily\small,
  keywordstyle=\color{blue!70!black},
  commentstyle=\color{gray}\itshape,
  stringstyle=\color{green!45!black},
  showstringspaces=false,
  numbers=none,
  frame=single,
  framesep=4pt,
  breaklines=true,
  columns=fullflexible,
  keepspaces=true,
}
\usepackage{subcaption}
\usepackage{pifont}
\newcommand{\cmark}{\textcolor{green!60!black}{\ding{51}}}  % ✓
\newcommand{\xmark}{\textcolor{red!70!black}{\ding{55}}}    % ✗
\usepackage{algorithm}
\usepackage{algorithmic}
\newcommand{\name}{{VAKRA}}
\title{\name{}: Evaluating Multi-Hop Reasoning Across APIs and Retrieval Under Tool-Use Policies}

\author{Author 1 \\ Address line \\  ... \\ Address line
        \And  ... \And
        Author n \\ Address line \\ ... \\ Address line}
\author{
 \textbf{Ankita Rajaram Naik},
 \textbf{Anupama Murthi},
 \textbf{Benjamin Elder},
 \textbf{Siyu Huo},
\\
 \textbf{Raavi Gupta},
 \textbf{Abhinav Jain},
 \textbf{Praveen Venkateswaran},
 \textbf{Abdulhamid Adebayo},
\\
 \textbf{Danish Contractor}
\\
IBM, Yorktown Heights, NY, USA
\\
 \small{
   \textbf{Correspondence:} \href{mailto:ankita.naik@ibm.com}{ankita.naik@ibm.com}, \href{mailto:anupama.murthi@ibm.com}{anupama.murthi@ibm.com}
 }
}

\begin{document}
\maketitle
\begin{abstract}
%\todo{good to include target audience - a benchmark for enterprise agent builders or so}

\textcolor{black}{Agents deployed in enterprise settings must reason across structured APIs and document collections, yet existing benchmarks evaluate these capabilities in isolation. We introduce \name{} (e\textbf{V}aluating \textbf{A}PI and \textbf{K}nowledge \textbf{R}etrieval \textbf{A}gents)\footnote{The name also alludes to the Sanskrit \emph{vakra}, meaning curved or indirect, evoking the non-linear reasoning paths agents must navigate.}, a benchmark of over $8{,}000$ executable APIs across $62$ domains with tasks spanning three settings of increasing difficulty: diverse API interaction styles, multi-hop reasoning over structured APIs, and multi-source reasoning with natural-language tool-use policy constraints. Correctness is verified by re-executing predicted tool calls against live APIs, accommodating multiple valid paths. Using a fixed ReAct harness to isolate model capabilities from agent architecture, we evaluate frontier and open-weight models and find that even the best model achieves only 70.4\% on single-hop endpoint-style tasks and drops to 50--51\% on compositional APIs; performance degrades by over 50\% as reasoning depth increases, and policy-constrained questions expose severe failures (as low as 2.4\% on unanswerable queries). Trace analysis shows failures concentrate at language-mediated reasoning - entity disambiguation, cross-source grounding, rather than tool invocation mechanics. Code is available https://github.com/IBM/VAKRA. Dataset is available https://huggingface.co/datasets/ibm-research/VAKRA}
\end{abstract}

\section{Introduction}

% \begin{figure}[t]
% \centering
% \includegraphics[width=\columnwidth]{figures/enterprise_example.png}
% \caption{Example enterprise workflow illustrating reasoning challenges including (i) \textbf{API-driven disambiguation}, (ii) \textbf{cross-source grounding}, (iii) \textbf{parameter alignment}, and (iv) \textbf{policy reasoning}.}
% \label{fig:ex1}
% \end{figure}

\begin{figure*}
\centering
\includegraphics[scale=0.32]{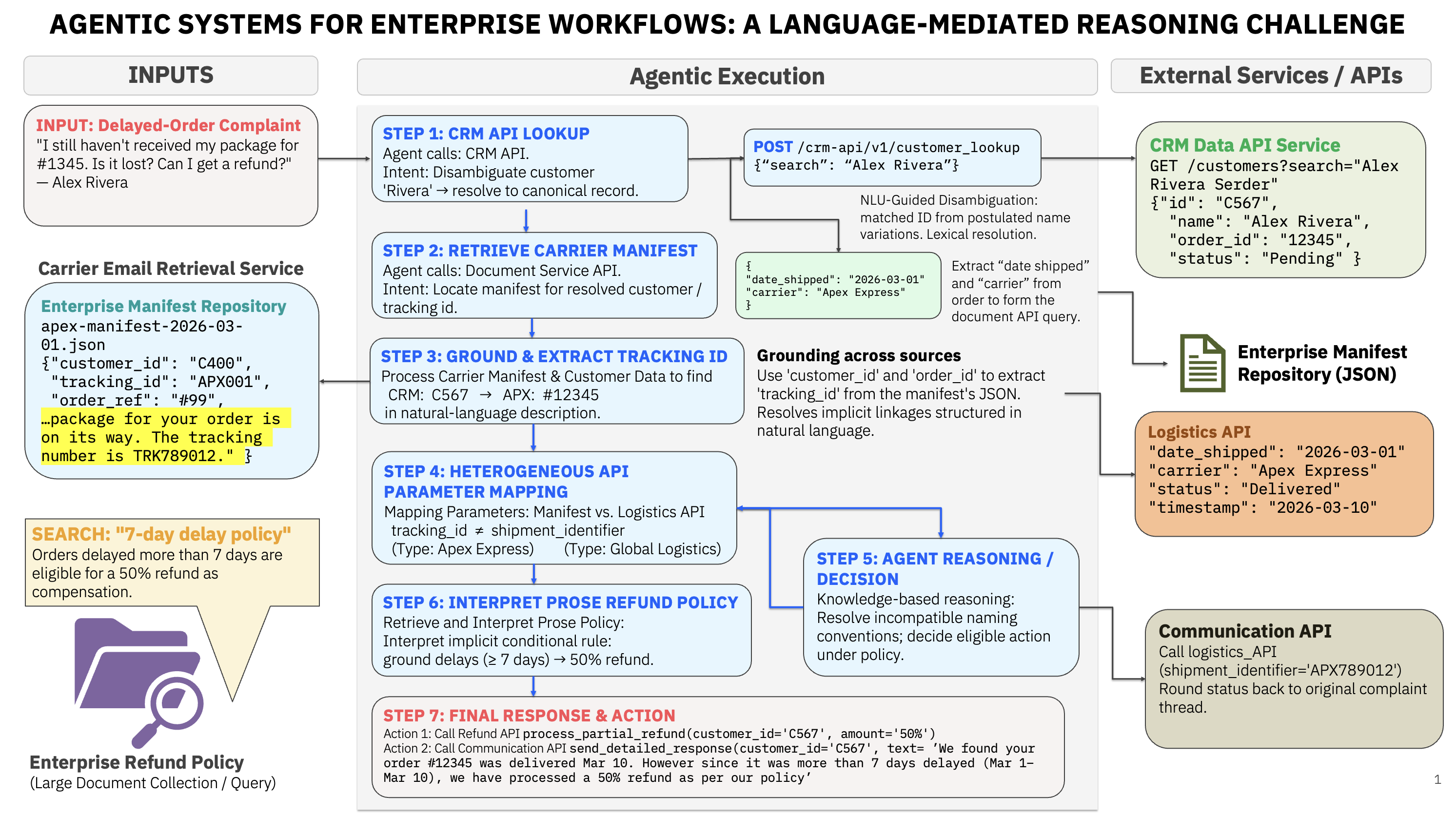}
\caption{Example enterprise workflow illustrating reasoning challenges including (i) \textbf{API-driven disambiguation}, (ii) \textbf{cross-source grounding}, (iii) \textbf{parameter alignment}, and (iv) \textbf{policy reasoning}.}
\label{fig:ex1}
\end{figure*}

Consider an agent resolving a delayed-order complaint as illustrated in Figure~\ref{fig:ex1}. The agent must disambiguate a customer record in a CRM (Step~$1$), extract a tracking identifier from carrier documentation (Step~$2$), map it to a logistics API parameter despite incompatible naming conventions (Steps~$3$--$5$), and interpret a refund policy to determine compensation (Steps~$6$--$7$). Each step requires language understanding: resolving entities under lexical mismatch, grounding information across heterogeneous sources, and interpreting constraints expressed in natural language. As agents are increasingly deployed across customer support, business intelligence, and financial operations~\cite{pan2026measuring}, reliable reasoning across APIs and document collections becomes a critical challenge.

Existing benchmarks only partially capture this setting. Prior work evaluates capabilities such as tool invocation~\citep{qin2023toolllmfacilitatinglargelanguage,bfcl}, web navigation~\citep{zhou2024webarenarealisticwebenvironment, boisvert2024workarena}, multi-turn dialogue~\citep{mtrag}, multi-hop document search~\cite{tang2024multihoprag, bfcl}, and policy adherence~\citep{zwerdling-etal-2025-towards, yao2025taubench, sopbench}. However, these capabilities are typically studied in isolation, leaving the compounding challenges of multi-hop reasoning across heterogeneous systems underexplored. This gap has practical consequences: a recent survey finds that $75\%$ of teams deploying agents evaluate them without formal benchmarks~\cite{pan2026measuring}.

In this paper, we introduce \textbf{\name} (e\textbf{V}aluating \textbf{A}PI and \textbf{K}nowledge \textbf{R}etrieval \textbf{A}gents), a benchmark for evaluating multi-hop reasoning across APIs and retrieval under tool-use policies in executable environments. \name{} builds on the structured API generation pipeline of \citet{elder2026liveapibench2500live}, and we transform BIRD-SQL~\citep{bird} into more than $8{,}000$ executable APIs derived from real databases across $62$ domains. We extend this foundation with domain-aligned document collections, \textcolor{black}{$2$--$5$ hop reasoning chains that require agents to combine API interaction with document retrieval {\em within a single trajectory}}, and a trajectory-level evaluation framework that re-executes predicted tool calls against live APIs\textcolor{black}{, accommodating multiple valid paths}. Tasks are organized into three settings of increasing difficulty: \textcolor{black}{(a)} diverse API styles, \textcolor{black}{(b)} multi-hop reasoning over structured APIs, and \textcolor{black}{(c)} multi-source reasoning with natural-language tool-use \textcolor{black}{policies}.

\textcolor{black}{Using a fixed ReAct harness to isolate model reasoning from architecture choices, we evaluate frontier and open-weight models. The strongest model (GPT-5.5) achieves 70.4\% on single-hop endpoint-style tasks but only 50--51\% on compositional business-intelligence APIs; most models lose over 50\% accuracy as reasoning depth increases, and policy-constrained questions prove especially difficult, with accuracy on unanswerable queries falling to 2.4\%. Trace analysis reveals that failures concentrate at language-mediated steps---entity disambiguation, cross-source grounding, and schema alignment---rather than tool invocation itself, indicating that compositional reasoning across heterogeneous sources remains the key bottleneck.}

\noindent \textbf{Contributions:}
(i) \textbf{Tool-grounded benchmark:} \textcolor{black}{Over $8{,}000$ executable APIs across $62$ domains, paired with domain-aligned document collections and a trajectory-level evaluation that re-executes tool calls against live APIs to verify correctness.}
(ii) \textbf{Compositional multi-hop tasks:} \textcolor{black}{$2$--$5$ hop chains combining structured API calls, cross-source retrieval, and natural-language tool-use policies---compounding challenges that prior benchmarks address only in isolation.}
(iii) \textbf{Empirical analysis:} \textcolor{black}{Model rankings invert across API interaction paradigms, performance degrades sharply with reasoning depth, and tool-use policy adherence remains a critical weakness with failures in localizing to language-mediated reasoning rather than tool invocation mechanics.}

\begin{table*}[ht]
\centering
\setlength{\tabcolsep}{1mm}
\resizebox{\textwidth}{!}{%
\begin{tabular}{l ccc ccccc c}
\toprule
\textbf{Benchmarks} &
\multicolumn{3}{c}{\textbf{Deployment}} &
\multicolumn{5}{c}{\textbf{Advanced Reasoning}} &
\multicolumn{1}{c}{\textbf{Evaluation}} \\
\cmidrule(lr){2-4} \cmidrule(lr){5-9} \cmidrule(lr){10-10}
& \textbf{Locally-Hosted} & \textbf{Real-World} & \textbf{Executable}
& \textbf{Nested API} & \textbf{Retriever} & \textbf{Multi-Hop} & \textbf{RAG+API} & \textbf{Policy}
& \textbf{Agentic/} \\
& \textbf{DB-Backed APIs} & \textbf{Data} & \textbf{at Eval Time}
& \textbf{Sequences} & \textbf{APIs} & \textbf{Queries} & \textbf{Joint Reasoning} & \textbf{Adherence}
& \textbf{Alt.\ Traces} \\
\midrule
%\textbf{NesTools}~\citep{han2024nestools}                             & \xmark & \xmark & \xmark & \cmark & \xmark & \xmark & \xmark & \xmark & \xmark \\
%\textbf{ToolAce}~\citep{liu2024toolace}                               & \xmark & \xmark & \xmark & \cmark & \xmark & \xmark & \xmark & \xmark & \xmark \\
%\textbf{API-Bank}~\citep{li2023api}                                   & \xmark & \xmark & \cmark & \xmark & \xmark & \xmark & \xmark & \xmark & \xmark \\
%\textbf{APIBench}~\citep{patil2024gorilla}                            & \xmark & \xmark & \cmark & \xmark & \xmark & \xmark & \xmark & \xmark & \xmark \\
\textbf{ToolBench}~\citep{xu2023toolmanipulationcapabilityopensource} & \xmark & \xmark & \cmark & \cmark & \xmark & \xmark & \xmark & \xmark & \xmark \\
\textbf{RestBench}~\citep{song2023restgpt}                            & \xmark & \cmark & \cmark & \cmark & \xmark & \xmark & \xmark & \xmark & \xmark \\
%\textbf{ToolQA}~\citep{zhuang2023toolqa}                              & \xmark & \cmark & \xmark & \xmark & \cmark & \xmark & \xmark & \xmark & \xmark \\
%\textbf{ToolAlpaca}~\citep{tang2023toolalpacageneralizedtoollearning} & \xmark & \xmark & \xmark & \xmark & \xmark & \xmark & \xmark & \xmark & \xmark \\
\textbf{ToolLLM}~\citep{qin2024toolllm}                               & \xmark & \cmark & \cmark & \xmark & \xmark & \xmark & \xmark & \xmark & \xmark \\
%\textbf{APIGen}~\citep{NEURIPS2024_61cce86d}                          & \xmark & \xmark & \xmark & \xmark & \xmark & \xmark & \xmark & \xmark & \xmark \\
\textbf{NESTFUL}~\citep{basu2025nestfulbenchmarkevaluatingllms}       & \xmark & \xmark & \xmark & \cmark & \xmark & \xmark & \xmark & \xmark & \xmark \\
%\textbf{BFCL V2}~\citep{bfcl}                                         & \xmark & \cmark & \cmark & \cmark & \xmark & \xmark & \xmark & \xmark & \xmark \\
\textbf{BFCL V2, V3}~\citep{bfcl}                                         & \xmark & \cmark & \cmark & \cmark & \xmark & \xmark & \xmark & \xmark & \xmark \\
\textbf{BFCL V4}~\citep{bfcl}                                         & \xmark & \cmark & \cmark & \cmark & \cmark & \cmark & \xmark & \xmark & \xmark \\
\textbf{WebArena}~\citep{zhou2024webarenarealisticwebenvironment}     & \xmark & \cmark & \cmark & \xmark & \xmark & \cmark & \xmark & \xmark & \xmark \\
\textbf{WorkArena}~\citep{boisvert2024workarena}                      & \xmark & \cmark & \cmark & \xmark & \xmark & \cmark & \xmark & \xmark & \xmark \\
\textbf{$\tau$-bench}~\citep{yao2025taubench}                         & \xmark & \cmark & \cmark & \xmark & \xmark & \cmark & \xmark & \cmark & \cmark \\
\textbf{AgentBench}~\citep{agentbench}                                & \xmark & \cmark & \cmark & \xmark & \xmark & \cmark & \xmark & \xmark & \cmark \\
\textbf{GAIA}~\citep{mialon2024gaia}                                  & \xmark & \cmark & \xmark & \xmark & \cmark & \cmark & \xmark & \xmark & \xmark \\
\textbf{ToolHop}~\citep{ye2025toolhop}                                & \xmark & \xmark & \cmark & \cmark & \xmark & \cmark & \xmark & \xmark & \xmark \\
\textbf{MCP-Bench}~\citep{wang2025mcp}                                & \xmark & \xmark & \cmark & \xmark & \xmark & \xmark  & \xmark & \xmark & \cmark \\
\textbf{LiveAPIBench}~\citep{elder2026liveapibench2500live}           & \cmark & \cmark & \cmark & \cmark & \xmark & \xmark & \xmark & \xmark & \cmark \\
\midrule
\textbf{\name (Ours)}                                               & \cmark & \cmark & \cmark & \cmark & \cmark & \cmark & \cmark & \cmark & \cmark \\
\bottomrule
\end{tabular}%
}
\caption{Comparison of tool-calling benchmarks across \emph{Deployment}, \emph{Advanced Reasoning}, and \emph{Evaluation} dimensions. Except for LiveAPIBench, none provide locally-hosted APIs backed by real databases (and document collections), and none prior to \name{} combine this with the full set of advanced reasoning and evaluation properties.}% \name uniquely combines nested API sequences, retriever-based APIs, multi-hop queries, RAG+API joint reasoning, policy adherence, and trace-level agentic evaluation accommodating alternative execution paths.}
\label{tab:benchmark_comparison}
%\caption{Comparison of tool-calling benchmarks across \emph{Deployment}, \emph{Advanced Reasoning}, and \emph{Evaluation} dimensions. Few benchmarks provide self-hosted APIs backed by real databases, and none prior to \name combine this with the full set of advanced reasoning and evaluation properties. }%\name uniquely combines nested API sequences, retriever-based APIs, multi-hop queries, RAG+API joint reasoning, policy adherence, and trace-level agentic evaluation accommodating alternative execution paths.}
\end{table*}

\section{Related Work} \label{sec:related}

We situate \name{} along three dimensions - executable grounding, cross-source compositional reasoning, and trajectory-level evaluation and focus on benchmarks most relevant to tool-grounded agent reasoning (Table~\ref{tab:benchmark_comparison}). We highlight three columns in the table that could be conflated: Nested API Sequences tests parameter-level compositional chaining where one call's output directly parameterizes the next; Multi-Hop Queries requires broader sequential reasoning including disambiguating entities, interpreting intermediate results, or selecting among actions, where earlier results inform but need not directly parameterize later calls; and RAG+API Joint Reasoning additionally requires retrieval outputs from an independent document collection to condition structured API parameters (or vice versa) within a single reasoning chain.

\noindent \textbf{Real executable grounding:}
Most tool-calling benchmarks either operate in synthetic settings~\citep{han2024nestools, liu2024toolace, li2023api}, rely on external APIs whose behavior changes over time~\citep{xu2023toolmanipulationcapabilityopensource, qin2024toolllm, guo-etal-2024-stabletoolbench}, or simulate execution via model-generated feedback~\citep{zhang2026geckosimulationenvironmentstateful}. Like \citet{elder2026liveapibench2500live}, \name{} self-hosts all APIs locally against real BIRD-SQL databases~\citep{bird} and document collections, ensuring deterministic, verifiable responses at evaluation time (see Table~\ref{tab:benchmark_comparison}).

\noindent \textbf{Cross-source and compositional reasoning:}
WebArena and WorkArena~\citep{zhou2024webarenarealisticwebenvironment, boisvert2024workarena} evaluate UI interaction and system state rather than data-level reasoning over schemas and documents. GAIA~\citep{mialon2024gaia} and AppWorld~\citep{trivedi-etal-2024-appworld} assess long-horizon planning but do not target cross-source~\footnote{GAIA tasks may incidentally combine file reading and web tool use, but are not designed to require retrieval outputs to directly condition structured API parameters within a single reasoning chain, or vice versa.}  grounding challenges such as entity resolution under lexical mismatch or alignment of retrieved content with structured API schemas. ToolQA~\citep{zhuang2023toolqa} provides retrieval and tabular query tools within a unified interface, but does not require agents to reconcile independently designed systems with heterogeneous schema conventions. \name{} explicitly targets this setting, requiring agents to interleave structured API calls with unstructured document retrieval as intermediate steps whose outputs directly parameterize subsequent tool calls within a single multi-hop reasoning chain.

\noindent \textbf{Policy-constrained workflows and trace-level evaluation:}
$\tau$-bench~\citep{yao2025taubench} is the closest benchmark to \name{} in spirit, but confines policy reasoning to narrow conversational domains without grounding decisions across independent data sources. \name{} has $62$ domains and requires agents to interpret and satisfy tool-use constraints expressed in natural language across multi-hop API and retrieval actions. Further, most benchmarks score final answers, individual call accuracy~\citep{bfcl}, or binary task success~\citep{zhou2024webarenarealisticwebenvironment, trivedi-etal-2024-appworld} — metrics that cannot distinguish whether intermediate steps such as entity resolution or policy interpretation were valid. \name{} instead verifies complete trajectories by re-invoking tool calls against live APIs, accommodating multiple valid execution paths and enabling targeted analysis of failures in language-mediated reasoning rather than just tool execution mechanics.

\section{Dataset} \label{sec:dataset}
We organize tasks into three settings of increasing difficulty (see a data sample in Appendix \ref{app:data_sample}).

\paragraph{(i) API Styles.} \label{sec:api-styles}
Following \citet{elder2026liveapibench2500live}, we instantiate three interaction styles that expose the underlying data through different interface abstractions. The first two are motivated by Business Intelligence (BI) APIs, while the third reflects reusable workflow and dashboard-oriented endpoints. We refer to these as SLOT, SEL, and Dashboard APIs respectively in our evaluation.
\textbf{Compositional interfaces (referred to as {\em SLOT})} provide a minimal set of general-purpose operations (e.g., filtering, aggregation, transformation) that must be composed into multi-step chains, requiring explicit planning over intermediate states. The agent needs to select between 9 general purpose tools provided.
\textbf{Expanded function interfaces (referred to as {\em SEL})} materialize parameterizations of generic functions as distinct tools, reducing per-call complexity while increasing the burden of selecting the correct operation from a larger candidate set. The agent needs to select from 26 available tools.
\textbf{Endpoint-style interfaces (referred to as {\em Dashboard} APIs}) provide highly specific, query-aligned endpoints that encapsulate most computation, shifting the challenge toward accurate query interpretation and endpoint selection. This task has an available tool list of 116 per sample (detailed tool distribution by domain in Appendix \ref{app:tool-distribution}).

\paragraph{(ii) Multi-hop Reasoning.} \label{sec:multi-hop-api-only}
We construct tasks requiring $2$--$5$ step reasoning chains over endpoint-style APIs (sample distribution by hops is present in Appendix \ref{app:data_statistics}), where outputs of earlier calls determine inputs to subsequent steps. These tasks require entity disambiguation, parameter extraction from intermediate responses, and API schema alignment across heterogeneous endpoints.

\paragraph{(iii) Multi-hop Multi-Source Reasoning.} \label{sec:multi-hop-multi-source}
We extend the previous setting to tasks requiring reasoning across structured APIs and unstructured document collections, in both single- and multi-turn settings. Agents must perform cross-source grounding aligning information extracted from text with API parameters, or using API outputs to inform retrieval while reconciling inconsistencies in naming and format. Beyond chaining, agents must decide \emph{when} to retrieve, \emph{what} to extract, and \emph{how} to integrate retrieved content into subsequent tool calls.
We augment a subset of these tasks with natural-language \textbf{tool-use policies} that govern \emph{source selection}, specifying which tools or retrieval collections are permissible for a given query.

Table~\ref{tab:dataset_stats} covers data statistics of tuning and test split of the benchmark. Of the 664 samples in multihop multisource dataset 244 samples have policies. Multi-source hop-type distribution for the remaining 420 queries is included in Appendix \ref{app:data_statistics}. Samples of the data are included in Appendix \ref{app:data_sample}.

\subsection{Dataset Construction} \label{sec:data-const}

\noindent\textbf{Tool Environment.}
We extend the API generation pipeline of \citet{elder2026liveapibench2500live}, which exposes over $8{,}000$ executable Python functions spanning $62$ BIRD-SQL~\citep{bird} domains, and enrich tool and argument descriptions following \citet{agarwal2025automated}. We supplement these with domain-specific retrieval tools: we populate ChromaDB indices with documents from ClapNQ~\citep{rosenthal2025clapnq} and Wikidata5M~\citep{wang2021kepler} to support document-grounded reasoning.

\noindent\textbf{Multi-hop Query Generation.}
Following \citet{trivedi2022musique}, we construct multi-hop tasks via a four-stage pipeline (detailed in Appendix~\ref{app:query_generation}): (i)~we  first extract named entities from BIRD-SQL queries and map them to Wikidata5M identifiers to build domain-specific knowledge graphs; (ii)~we then link APIs whose outputs parameterize other APIs' inputs into a query connectivity graph and traverse it depth-first to produce $1$--$3$ hop reasoning chains (sampling hop counts with weights $0.10$/$0.60$/$0.30$ for $1$/$2$/$3$ hops);  the queries associated with these chains are generated by merging the per-hop queries using an LLM; %\footnote{Due to the similar nature of queries and tools present in \citep{shlomov2026benchmarks} the data is adapted as a domain bpo in the multihop setting. This adds the expanded 4-5 hop queries.}; 
(iii)~we fetch and ground Wikipedia passages against the knowledge graph to introduce retrieval-augmented links, generating combined API+RAG questions using an LLM \footnote{Mistral-Large-2411~\cite{mistral_large_2411}}; (iv)~lastly, we generate retrieval-only multi-turn dialogues following \citet{lee2024multidocumentgroundedmultiturnsynthetic} and apply cross-source answerability filtering to ensure RAG questions cannot be answered via structured APIs and vice versa (See Appendix ~\ref{app:multi_rag_pipeline} for more details).

\noindent\textbf{Retrieval Index Construction.} \label{sec:retrieval-index-construction}
We construct domain-specific indices from Wikidata5M and ClapNQ documents. We apply an LLM-based filter \footnote{mistralai/Mixtral-8x22B-v0.1} that removes documents capable of answering any API query and discards RAG queries answerable via APIs, ensuring clean source separation. We then index the surviving documents per domain using ChromaDB.

\begin{table}[t]
\centering
\scriptsize
\setlength{\tabcolsep}{3pt}
\renewcommand{\arraystretch}{1.05}

\begin{tabular}{
>{\raggedright\arraybackslash}m{2.4cm}
>{\centering\arraybackslash}m{1.15cm}
>{\centering\arraybackslash}m{1.15cm}
>{\centering\arraybackslash}m{0.8cm}
>{\centering\arraybackslash}m{0.75cm}
}
\toprule

\textbf{Setting}
& \textbf{\# Domains}
& \textbf{\# Samples}
& \textbf{Avg.\ Tool Calls}
& \textbf{Max Tool Calls} \\

\midrule
\multicolumn{5}{c}{\textbf{Tuning Split}} \\
\midrule

BI APIs (SEL)
& 17 & 710 & 4.15 & 12 \\

BI APIs (SLOT)
& 16 & 614 & 3.9 & 9 \\

Dashboard APIs
& 40 & 1,860 & 1.00 & 1 \\

Multi-hop Reasoning
& 28 & 346 & 2.05 & 3 \\

\makecell[l]{Multi-source Multi-hop\\Reasoning}% with Policy}
& 36 & 898 & 1.05 & 3 \\

\midrule
\multicolumn{5}{c}{\textbf{Test Split}} \\
\midrule
BI APIs (SEL)
& 18 & 549 & 4.11 & 10 \\

BI APIs (SLOT)
& 33 & 1397 & 3.89 & 10 \\

Dashboard APIs
& 17 & 1,597 & 1.00 & 1 \\

Multi-hop Reasoning
& 38 & 869 & 2.04 & 5 \\

\makecell[l]{Multi-source Multi-hop\\Reasoning} % with Policy}
& 41 & 644 & 1.34 & 4 \\
\bottomrule

\end{tabular}
\caption{
Dataset statistics across tuning and test splits.
}
\vspace{-4ex}
\label{tab:dataset_stats}
\end{table}

% \begin{table}[t]
% \centering
% \scriptsize
% \setlength{\tabcolsep}{4pt}

% \begin{tabular}{>{\centering\arraybackslash}m{2.5cm}cccc}
% \toprule
%  & \textbf{\# Domains} & \textbf{ \# Samples} & \textbf{Avg \# tool calls} & \textbf{Max \# tool calls} \\
% \midrule

% \multicolumn{5}{c}{\textbf{Training Split}} \\
% \midrule

% BI APIs (SEL) &  &  &  &  \\
% BI APIs (SLOT) &  &  &  &  \\
% Dashboard APIs & 40 & 1,860 & 1.00 & 1 \\
% Multihop Reasoning & 28 & 346 & 2.05 & 3 \\

% MultiHop MultiSource Reasoning with Policy
% & 36 & 898 & 1.05 & 3 \\

% \midrule
% \multicolumn{5}{c}{\textbf{Test Split}} \\
% \midrule

% BI APIs (SEL) &  &  &  &  \\
% BI APIs (SLOT) &  &  &  &  \\
% Dashboard APIs & 17 & 1,597 & 1.00 & 1 \\
% Multihop Reasoning & 38 & 869 & 2.04 & 5 \\

% MultiHop MultiSource Reasoning with Policy
% & 41 & 644 & 1.34 & 4 \\

% \bottomrule
% \end{tabular}

% \caption{Dataset statistics across training and test splits for benchmark tasks.}
% \label{tab:dataset_stats}
% \end{table}

\subsection{Data Quality Assessment}
\label{sec:data_quality_assessment}

As prior work on LLM-based query generation and conversational retrieval has shown that automatically generated queries often suffer from hallucinated constraints, inconsistent reasoning chains, and retrieval shortcut artifacts~\citep{agentcq2024,llmenhancedquery2025,qaexpand2025}, we conducted a small human study for \textbf{Multi-hop Reasoning} and \textbf{Multi-hop Multi-Source Reasoning} settings to assess the quality of the LLM-generated compositional questions (See Appendix \ref{app:human_eval}). 

Following connected reasoning principles introduced in MuSiQue~\citep{trivedi2022musique}, we evaluate whether each generated question requires coherent multi-step reasoning rather than disconnected shortcut inference. We additionally adopt evaluation dimensions motivated by prior work on RAG evaluation~\citep{es2024ragas}, multi-hop QA~\citep{yang2018hotpotqa,trivedi2022musique}, and NLG evaluation~\citep{papineni2002bleu,liu2023geval}.

We sample $60$ questions from each setting ($869$ Multi-hop Reasoning, $644$ Multi-hop Multi-Source Reasoning) via stratified sampling: the $62$ domains are grouped into $12$ semantic clusters (Appendix~\ref{tab:domain-clusters}), and we draw $5$ queries per cluster. Three annotators independently score each question across $5$ rubric dimensions—faithfulness, logical consistency, answer leakage, context sufficiency, and cross-source entity consistency—using an ordinal scale.~\footnote{$120$ questions $\times$ $3$ annotators $\times$ $5$ dimensions $=$ $1{,}800$ total judgments.}
% For \textbf{Multi-hop Reasoning} questions, annotators evaluate:
% (i)~\textbf{Faithfulness}, measuring whether the merged question introduces unsupported entities or relations;
% (ii)~\textbf{Naturalness}, measuring grammaticality and whether the merged question reads like a realistic human query;
% (iii)~\textbf{Logical Consistency}, measuring whether the composed question contains contradictory or logically incompatible constraints; 
% (iv)~\textbf{Answer Leakage}, measuring whether the generated question trivially reveals intermediate-hop or final answers; and
% (v)~\textbf{Context Sufficiency}, measuring whether the question contains enough information to execute the required reasoning chain.

% For \textbf{Multi-hop Multi-Source Reasoning} questions, we additionally evaluate:
% (i)~\textbf{Retrieval Sufficiency Score}, measuring whether the retrieved documents contain sufficient evidence to answer the RAG component of the query; and
% (ii)~\textbf{Cross-Hop Entity Consistency Score}, measuring whether entities propagated across API and retrieval hops are correctly inferred and grounded in retrieved evidence.

Each dimension is evaluated on a $1$--$4$ ordinal scale, where lower scores correspond to severe failures and higher scores correspond to fully grounded and coherent reasoning structure. \textcolor{black}{A sample is considered high-quality if its average score across all metrics is $\geq 3.0$.}

\noindent{\bf Results:} Inter-annotator agreement is 77\% for Multi-hop Reasoning and 90\% for Multi-hop Multi-Source Reasoning. Under the quality threshold, 87\% of Multi-hop Reasoning samples and 96\% of Multi-hop Multi-Source Reasoning samples are rated as high-quality. See Appendix \ref{app:human_eval} for more details about the Human Study.

\begin{table*}[ht]
\centering
\scriptsize
\setlength{\tabcolsep}{2.5pt}
\begin{tabular}{@{}m{0.32cm}@{\hspace{2pt}}lccccccc|cccc@{}}
\toprule
&
\multirow{2}{*}{\textbf{Model}} &
\multicolumn{3}{c}{\textbf{API Styles}} &
\multirow{2}{*}{\shortstack{\textbf{Multi-hop}\\\textbf{Reasoning}}} &
\multirow{2}{*}{\shortstack{\textbf{Multi-Source}\\\textbf{Multi-hop}\\\textbf{ }}} &
\multirow{2}{*}{\shortstack{\textbf{Avg.}\\\textbf{Score}}} &
\multirow{2}{*}{\shortstack{\textbf{Avg.}\\\textbf{Tool Calls}}} &
\multicolumn{3}{|c}{\textbf{Policy Category Success Breakdown (\%)}} \\
\cmidrule(lr){3-5}
\cmidrule(lr){10-12}
&
& \shortstack{\textbf{BI APIs}\\\textbf{(SEL)}}
& \shortstack{\textbf{BI APIs}\\\textbf{(SLOT)}}
& \shortstack{\textbf{Dashboard}\\\textbf{APIs}}
& & & &
& \shortstack{\textbf{Policy Updates}\\\textbf{Answer}}
& \shortstack{\textbf{No Effect}\\\textbf{on Answer}}
& \shortstack{\textbf{No}\\\textbf{Policy}} \\
\midrule

\raisebox{-0.18\height}{\includegraphics[height=0.27cm]{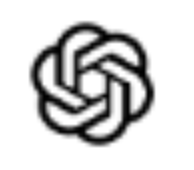}}
& GPT-5.5
& \textbf{51.0} & \textbf{50.04} & \textbf{70.4} & \textbf{52.4} & \textbf{26.0}
& \textbf{50.1} & 3.7
& 4.9 & 40.7 & 31.9 \\

\raisebox{-0.18\height}{\includegraphics[height=0.27cm]{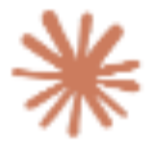}}
& Claude-Opus-4.7
& --- & --- & --- & 43.4 & 18.4
& --- & ---
& 2.4 & 31.5 & 21.5 \\

\raisebox{-0.18\height}{\includegraphics[height=0.27cm]{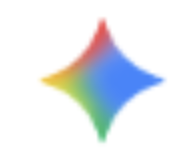}}
& Gemini-3-Flash-Preview
& 38.6 & 39.0 & 60.3 & 36.9 & 16.7
& 38.7 & 4.1
& 2.4 & 25.9 & 21.8  \\

\raisebox{-0.18\height}{\includegraphics[height=0.27cm]{figures/logos/claude.png}}
& Claude-Sonnet-4.5
& 35.3 & 38.4 & 49.5 & 38.1 & 17.3
& 35.9 & 3.0
& 3.7 & 27.2 & 21.2 \\

\raisebox{-0.18\height}{\includegraphics[height=0.27cm]{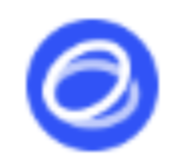}}
& GLM-5.1
& 35.7 & 39.3 & 47.5 & 32.7 & 14.5
& 34.4 & 3.7
& 7.3 & 24.7 & 16.1 \\

\midrule
\raisebox{-0.18\height}{\includegraphics[height=0.27cm]{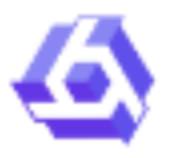}}
& Qwen-3.5-397B
& 42.3 & 47.8 & 46.7 & 30.0 & 16.6
& 37.1 & 5.5
& 4.9 & 25.9 & 17.4 \\

\raisebox{-0.18\height}{\includegraphics[height=0.27cm]{figures/logos/openai.png}}
& GPT-OSS-120B
& 40.1 & 42.7 & 50.5 & 25.1 & 15.5
& 35.0 & 2.6
& 17.1 & 24.7 & 16.8 \\

\raisebox{-0.18\height}{\includegraphics[height=0.27cm]{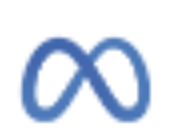}}
& LLaMA-405B
& 29.0 & 39.7 & 54.4 & 26.8 & 12.8
& 32.6 & 1.7
& 3.7 & 17.9 & 15.0 \\

% \raisebox{-0.18\height}{\includegraphics[height=0.27cm]{figures/logos/mistral.png}}
% & Mistral-Large-3-675B
% & 31.7 & 33.9 & 41.6 & 23.6 & 12.5
% & 29.1 & 1.9
% & 4.9 & 19.8 & 14.5 \\

\midrule
\raisebox{-0.18\height}{\includegraphics[height=0.27cm]{figures/logos/qwen.png}}
& Qwen-2.5-72B-Instruct
& 34.8 & 40.4 & 50.2 & 20.3 & 11.4
& 31.7 & 2.5
& 3.7 & 16.7 & 14.5 \\

\raisebox{-0.18\height}{\includegraphics[height=0.27cm]{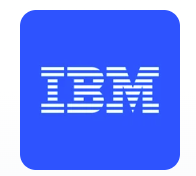}}
& Granite-4h-small
& 28.1 & 26.1 & 50.0 & 26.2 & 12.4
& 29.1 & 2.3
& 3.7 & 21.6 & 15.8 \\

\raisebox{-0.18\height}{\includegraphics[height=0.27cm]{figures/logos/meta.png}}
& LLaMA-3.3-70B-Instruct
& 27.7 & 33.9 & 49.2 & 13.2 & 13.7
& 27.7 & 1.7
& 4.9 & 18.5 & 16.3 \\

\bottomrule
\end{tabular}

\caption{
Agent completion rates across task settings:
API Styles (§\ref{sec:api-styles}),
Multi-hop Reasoning (§\ref{sec:multi-hop-api-only}),
and Multi-Source Multi-hop reasoning with policy adherence
(§\ref{sec:multi-hop-multi-source}). Number of tool calls per model per setting is also provided in Table ~\ref{tab:avg-predicted-tool-calls-combined}. The final three columns report success percentages across policy categories in the multihop multisource with policy adherence evaluation setting. Claude Opus 4.7 was evaluated on a subset due to cost considerations.}

\label{tab:main_results}
\end{table*}

\section{Evaluation}
\label{sec:evaluation}

We evaluate agent outputs via a waterfall mechanism: each stage gates the next, so failures are attributed to the earliest point of breakdown.

\noindent{\bf Stage 1: Tool-Sequence Verification:}
Because agents may solve a task via alternative valid tool sequences, we do not enforce strict step-level matching. Instead, we re-execute each predicted tool call against the live environment and compare the \emph{set of tool responses} against the ground truth. We first apply a programmatic containment check verifying whether all ground-truth information is recovered. For inconclusive cases (partial matches, semantic equivalence, formatting differences), we apply an LLM-based judge adapted from the CRAG framework ~\citep{yang2024crag} to determine whether the predicted trajectory retrieves all required information.

\noindent{\bf Stage 2: Final Response Evaluation:}
Only trajectories passing Stage~1 proceed. An LLM judge evaluates the agent’s final response for (i)~groundedness in tool responses and (ii) the answer correctness framework from \citet{es2024ragas} used to calculate factual consistency with the ground-truth answer, allowing phrasing variation (LLM-as-Judge prompts included in Appendix \ref{app:eval_details}).

\noindent{\bf Stage 3: Policy Adherence:}
For policy-constrained tasks, we verify deterministically that no disallowed sources were consulted, independent of answer correctness.

Following prior work establishing GPT-family models as reliable LLM-as-Judges~\cite{yang2024crag,zheng2023judging,liu2024evaluating,li2024crowdsourced} and recent evidence that reasoning-oriented open models can achieve performance competitive with proprietary GPT-based evaluators on RewardBench-style benchmarks~\cite{together2025judge,wang2026reasoningjudge}, we employ GPT-OSS-120B~\cite{openai2025gptoss120bgptoss20bmodel} as the judge model for both Stage~1 and Stage~2 with temperature~$0$.

\noindent{\bf Scoring:}
The final score averages across all task settings; within multi-source tasks, multi-source queries receive double weight relative to single-source queries. %We use GPT-OSS-120B as the LLM judge with temperature~$0$.

\subsection{Execution Environment and Harness}
\paragraph{Self-Hosted Infrastructure.}
All tools ship in a \emph{single Docker image} (\texttt{benchmark\_environ}),
instantiated as one container per capability. The environment hosts
\emph{structured API tools} (which are powered by underlying SQL queries over domain-specific SQLite databases)
and \emph{retrieval tools} (semantic search over 62 ChromaDB collections
embedded with IBM \texttt{granite-embedding-english-r2}); structured tools are
surfaced as slot-filling/selection interfaces for SEL/SLOT and as REST endpoints
otherwise. %Routing uses zero-overhead process replacement: a shared entrypoint
% reads \texttt{CAPABILITY\_ID} and \texttt{os.execv()}s into the right MCP server.
% REST-backed servers need no hand-written tool definitions---each converts its
% FastAPI OpenAPI spec into typed MCP tools and filters to the active domain.
Agents communicate over the Model Context Protocol on stdio; databases and raw indices are never exposed to agents. %, so every response is deterministic, verifiable, and
% free of external dependencies.

% \todo{DC: I didn't understand this para}\paragraph{Adversarial Tool Surface.}
% In \name{}, selecting the \emph{right} tool is part of the task: the
% environment deliberately seeds the tool surface with plausible distractors. The
% capability-4 server applies \emph{asymmetric} filtering---structured tools are
% restricted to the primary domain, while retrieval tools also expose confusable
% ``negative'' domains. Combined with domains exposing up to several hundred
% tools, this makes tool grounding a first-class axis of difficulty.

% {\todo: Appendix stuff}\paragraph{Reproducible, One-Command Setup.}
% The environment needs no hosted service, API key, or cloud dependency: the image
% and backing data are published to HuggingFace, and \texttt{docker compose up -d}
% launches all four containers. To guard against tool-surface drift, we commit a
% SHA-256 checksum over tool names and input schemas for each
% \texttt{(capability, domain)} pair; both server and runner verify it before any
% query, and a mismatch raises a hard error---keeping every reported number
% reproducible against a known tool surface.

\noindent{\bf Benchmark Runner and Agent Implementation:}
A runner orchestrates the evaluation lifecycle---loading benchmark items,
connecting to each task container's MCP server via \texttt{stdio\_client},
streaming verified tool definitions, and recording a complete trajectory (tool
calls, responses, final answer) to structured JSON under a fixed per-query
timeout. Every model is wrapped in a LangGraph ReAct agent behind a uniform
\texttt{AgentInterface}, so open and closed models from a provider-agnostic
factory (Anthropic, OpenAI, Ollama, LiteLLM, watsonx) are evaluated identically.
MCP tools are converted to typed LangChain \texttt{StructuredTool}s exposing
complete parameter signatures. The agent is never told task
characteristics i.e, the hops required, or whether retrieval is needed and it
answers given only its tools and tool-use policy, if present
(see Appendix~\ref{app:harness_details}).

\paragraph{Why \textsc{ReAct}?}
We adopt the \textsc{ReAct} paradigm~\cite{yao2023react} for three
reasons. \emph{(i)~Minimal and model-agnostic:} it is a bare
reason--act--observe loop with no built-in planner or task
decomposition, so benchmark scores reflect the underlying model's
capability rather than harness engineering, providing a fair common
baseline across models. \emph{(ii)~Suited to large tool spaces:} the
explicit reasoning step forces the model to articulate \emph{why} a tool
is selected and \emph{how} its arguments are grounded, which improves
tool selection and yields interpretable traces---valuable when domains
expose up to $328$ tools (Appendix~\ref{app:tool-distribution}).
\emph{(iii)~Self-correcting:} each tool response (an error, an empty
result, a malformed output) is fed back as an observation, letting the
agent revise its action within the same episode rather than failing
silently.
%\noindent\textbf{Judges and Metrics}

\begin{table*}[t]
\centering
\footnotesize
% \setlength{\tabcolsep}{4pt}
% \renewcommand{\arraystretch}{1.1}
% \begin{tabular*}{\textwidth{@{\extracolsep{\fill}}l|rrrr|rrrr|rrrr@{}}
\begin{tabular}{l|rrrr|rrrr|rrrr}
\toprule

& \multicolumn{4}{c|}{\textbf{Dashboard APIs}} 
& \multicolumn{4}{c|}{\textbf{BI APIs (SEL)}} 
& \multicolumn{4}{c}{\textbf{BI APIs (SLOT)}} \\

\cmidrule(lr){2-5}
\cmidrule(lr){6-9}
\cmidrule(lr){10-13}

\textbf{Model}
& Tool & ArgN & ArgV & Gnd
& Tool & ArgN & ArgV & Gnd
& Tool & ArgN & ArgV & Gnd \\

\midrule
\raisebox{-0.18\height}{\includegraphics[height=0.27cm]{figures/logos/openai.png}}
GPT-5.5
& 92.8 & 92.6 & 81.8 & 70.4
& 60.7 & 60.1 & 58.8 & 51.0
& 86.6 & 83.3 & 64.1 & 50.0 \\

\raisebox{-0.18\height}{\includegraphics[height=0.27cm]{figures/logos/gemini.png}}
Gemini-3-Flash
& 91.5 & 91.4 & 81.8 & 60.3
& 46.8 & 45.7 & 44.1 & 38.6
& 69.0 & 47.7 & 47.1 & 39.0 \\

\raisebox{-0.18\height}{\includegraphics[height=0.27cm]{figures/logos/claude.png}}
Claude-Sonnet-4.5
& 84.8 & 84.6 & 76.3 & 49.5
& 49.4 & 47.9 & 45.7 & 35.3
& 82.5 & 51.8 & 49.5 & 38.4 \\

\raisebox{-0.18\height}{\includegraphics[height=0.27cm]{figures/logos/glm.png}}
GLM-5.1
& 83.9 & 83.8 & 73.3 & 47.5
& 48.5 & 47.4 & 46.4 & 35.7
& 85.1 & 53.3 & 52.5 & 39.3 \\

\midrule
\raisebox{-0.18\height}{\includegraphics[height=0.27cm]{figures/logos/qwen.png}}
Qwen-3.5-397B
& 81.5 & 81.3 & 71.4 & 46.7
& 58.8 & 55.9 & 55.9 & 42.3
& 91.8 & 88.0 & 72.2 & 47.8 \\

\raisebox{-0.18\height}{\includegraphics[height=0.27cm]{figures/logos/openai.png}}
GPT-OSS-120B
& 72.7 & 72.3 & 59.9 & 50.5
& 52.3 & 50.1 & 50.1 & 40.1
& 86.2 & 85.3 & 64.0 & 42.7 \\

\raisebox{-0.18\height}{\includegraphics[height=0.27cm]{figures/logos/meta.png}}
LLaMA-405B
& 83.2 & 82.7 & 60.8 & 54.4
& 37.9 & 34.8 & 33.5 & 29.0
& 78.4 & 47.0 & 43.8 & 39.7 \\

% \raisebox{-0.18\height}{\includegraphics[height=0.27cm]{figures/logos/mistral.png}}
% Mistral-Large-3-675B
% & 64.0 & 63.9 & 52.5 & 41.6
% & 41.3 & 38.6 & 38.6 & 31.7
% & 64.4 & 63.9 & 44.0 & 33.9 \\

\midrule
\raisebox{-0.18\height}{\includegraphics[height=0.27cm]{figures/logos/qwen.png}}
Qwen-2.5-72B-Instruct
& 75.5 & 75.3 & 59.5 & 50.2
& 43.0 & 41.0 & 41.0 & 34.8
& 78.7 & 77.0 & 51.5 & 40.4 \\

\raisebox{-0.18\height}{\includegraphics[height=0.27cm]{figures/logos/granite.png}}
Granite-4h-small
& 76.7 & 76.5 & 58.1 & 50.0
& 41.3 & 34.2 & 33.7 & 28.1
& 78.2 & 39.2 & 34.1 & 26.1 \\

\raisebox{-0.18\height}{\includegraphics[height=0.27cm]{figures/logos/meta.png}}
LLaMA-3.3-70B-Instruct
& 73.8 & 73.5 & 54.8 & 49.2
& 34.2 & 31.5 & 31.5 & 27.7
& 70.9 & 69.3 & 38.7 & 33.9 \\

\bottomrule
\end{tabular}

\caption{Sieve of success: percentage of instances surviving each evaluation stage across Dashboard APIs, SEL APIs, and SLOT APIs. Each stage is cumulative: Correct Tool Names (Tool) $\rightarrow$ Correct Argument Names (ArgN) $\rightarrow$ Correct Argument Values (ArgV) $\rightarrow$ Fully Correct Grounded Answers (Gnd).}
\label{tab:sieve_success_combined}
\end{table*}
%\subsection{Baselines}
\section{Results}%\todo{Suggested outline}

\textcolor{black}{Our experiments are organized around three research questions. \textbf{RQ1 (API Styles):} How do models perform across different API interaction paradigms (SLOT, SEL, Dashboard), and where in the tool-calling pipeline do failures concentrate? \textbf{RQ2 (Multi-hop Reasoning):} How does performance degrade as the number of reasoning hops increases? \textbf{RQ3 (Multi-Source Reasoning \& Policy Adherence):} How do models handle reasoning that combines structured APIs with document retrieval, and can they adhere to natural-language tool-use policies?} %We address RQ1 in \S\ref{sec:api-styles-results} and the sieve-of-success analysis, RQ2 in \S\ref{sec:multihop-results}, and RQ3 in \S\ref{sec:multisource-results}.}

\noindent We summarize the full set of results in Table \ref{tab:main_results}.

\subsection{API Styles} \label{sec:api-styles-results}

Consistent with findings in \cite{elder2026liveapibench2500live}, our experiments on API styles (Table \ref{tab:main_results}) demonstrate that all models find the BI APIs\footnote{In contrast to \citet{elder2026liveapibench2500live}, we support verification of alternate traces in our evaluation framework %twhich lead to an improved performance of BI API tasks.
} more challenging to work with as compared to Dashboard APIs. GPT5.5 performs the best on all API styles, and medium-scale models can sometimes outperform large-scale open source models (eg: Qwen2.5B-72B-Instruct).  Within the BI APIs collection, models usually tend to perform better on the SLOT subset than SEL - the SLOT has a smaller number of generic tools with a large number of parameter values to fill, while the SEL collection has a larger set of tools and fewer parameters per tool. It is also interesting to note that the relative order of performance for models within each subset (BI vs Dashboard) is very different (eg: the lowest-performing model on BI APIs Granite-4h-small is \textcolor{black}{better than} many large models on the Dashboard API tasks including Qwen3.5-397B, GLM-5.1).

\subsection{MultiHop and Multi-source Reasoning} \label{sec:multihop-results} \label{sec:multisource-results}

% \noindent{\bf API Styles:} Unsurprisingly, we find that all models find the BI APIs more challenging to work with as compared to Dashboard APIs. Within the BI APIs collection, models usually tend to perform better on the SEL collection than SLOT - the SLOT has a smaller number of generic tools with a large number of parameter values to fill, while the SEL collection has a larger set of tools and fewer parameters per tool. This focus is reflected in the relative errors \todo{Needs to be main paper} (Table \ref{fig:sel_slot_bar_chart} made by models using these two tool collections. In the SLOT collection, we find that most made a substantial number of errors producing incorrect names for the tool {\em arguments}. %This is largely the reason that GPT-OSS-120b performed so well overall in this segment of the benchmark.
% With fewer parameters to fill, the same models made very few such errors when using the SEL  collection, but they made many more errors selecting the correct tools, reflecting the increased difficulty of choosing from a larger (and dynamic) tool set.

% On the other hand, the primary challenge with dashboard queries is tool-selection (no sequencing) and the scores are much higher. However, it is interesting to note that the relative order of performance for models within each subset (BI vs Dashboard) is very different (eg: the lowest performing model on BI APIs Granite-4h-small is better the many large models on the Dashboard API tasks including Mistra-Large-3-675B, Qwen3.5-397B, GLM-5.1).
% \\
\noindent{\bf Multi-hop Reasoning.} Performance drops substantially from Dashboard APIs to multi-hop chains (Table~\ref{tab:main_results}): models must reason over intermediate tool outputs to identify subsequent tools and their parameterization. The degradation steepens with hop count—all models except GPT-5.5 lose over 50\% accuracy as chains lengthen (Appendix Figure~\ref{fig:multihop_reaosning_error_chart}).

\noindent{\bf Multi-hop Multi-Source Reasoning.} Adding retrieval steps further compounds difficulty, as reflected in the lower scores in
Table~\ref{tab:main_results} (which include policy-constrained instances). Closed models perform comparably, with the exception of GPT-5.5 which maintains a consistent lead.

\noindent{\bf Tool-use Policy.} To disentangle effects of cross-source hops from policy constraints, Table~\ref{tab:main_results} reports three sub-categories: policy makes the question unanswerable, policy has no effect, and no policy. Overall trends hold across sub-categories, but unanswerable questions expose sharp model-specific weaknesses: Claude Opus 4.7 scores 2.4\% (vs.\ 3.7\% for the next-lowest models), suggesting it fails to recognize when a policy renders a question unanswerable and instead forces an answer. GLM-5.1 struggles separately with processing long tool responses. %In contrast, GLM-5.1 achieves the highest score in this category (47.6\%), indicating a stronger ability to respect policy constraints.}

\begin{table}[t]
\centering
\scriptsize
\setlength{\tabcolsep}{4pt}
\renewcommand{\arraystretch}{1.05}

\begin{tabular}{
>{\raggedright\arraybackslash}p{2.4cm}
>{\centering\arraybackslash}p{0.9cm}
>{\centering\arraybackslash}p{1.1cm}
>{\centering\arraybackslash}p{0.9cm}
>{\centering\arraybackslash}p{1.1cm}
}
\toprule

& \multicolumn{2}{c}{\textbf{BI APIs}}
& \multicolumn{2}{c}{\textbf{Dashboard APIs}} \\

\cmidrule(lr){2-3}
\cmidrule(lr){4-5}

\textbf{Model}
& \textbf{Extraction Errors (\%)}
& \textbf{Hallucinate (\%)}
& \textbf{Extraction Errors (\%)}
& \textbf{Hallucinate (\%)} \\

\midrule
GPT-5.5 & 4.5 & 95.5 & 38.2 & 61.8 \\
Gemini3-Flash & 5.8 & 94.2 & 35.3 & 64.7 \\
Claude-Sonnet-4.5 & 8.3 & 91.7 & 29.3 & 70.7 \\
GLM-5.1 & 12.1 & 87.9 & 29.0 & 71.0 \\
\midrule
Qwen3.5-397B & 9.5 & 90.5 & 23.6 & 76.4 \\
LLaMA-405B & 17.1 & 82.9 & 45.6 & 54.4 \\
GPT-OSS-120B & 24.2 & 75.8 & 31.1 & 68.9 \\
% Mistral-Large3-675B & 13.5 & 86.5 & 27.7 & 72.3 \\
\midrule
Granite4-Small & 13.0 & 87.0 & 43.6 & 56.4 \\
LLaMA-3.3-70B-Inst & 7.1 & 92.9 & 29.2 & 70.8 \\
Qwen2.5-72B-Inst & 11.5 & 88.5 & 20.1 & 79.9 \\
\bottomrule
\end{tabular}

\vspace{-3ex}
\caption{
Percentage distribution of extraction errors and hallucinations among samples in the Gnd error bucket from Table~\ref{tab:sieve_success_combined}.
}

\label{tab:cap1-cap2-ragas-dialogue0-counts}
\end{table}

% \begin{table}[t]
% \centering
% \scriptsize
% \setlength{\tabcolsep}{4pt}
% \renewcommand{\arraystretch}{1.05}

% \begin{tabular}{
% >{\raggedright\arraybackslash}p{2.4cm}
% >{\centering\arraybackslash}p{0.9cm}
% >{\centering\arraybackslash}p{1.1cm}
% >{\centering\arraybackslash}p{0.9cm}
% >{\centering\arraybackslash}p{1.1cm}
% }
% \toprule

% & \multicolumn{2}{c}{\textbf{BI APIs}}
% & \multicolumn{2}{c}{\textbf{Dashboard APIs}} \\

% \cmidrule(lr){2-3}
% \cmidrule(lr){4-5}

% \textbf{Model}
% & \textbf{Extraction Errors}
% & \textbf{Hallucinate}
% & \textbf{Extraction Errors}
% & \textbf{Hallucinate} \\

% \midrule
% GPT-5.5 & 8 & 170 & 68 & 110 \\
% Gemini3-Flash & 8 & 131 & 121 & 222 \\
% Claude-Sonnet-4.5 & 17 & 189 & 125 & 302 \\
% GLM-5.1 & 29 & 211 & 119 & 292 \\
% \midrule
% Qwen3.5-397B & 25 & 237 & 93 & 301 \\
% LLaMA-405B & 14 & 68 & 47 & 56 \\
% GPT-OSS-120B & 48 & 150 & 47 & 104 \\
% Mistral-Large3-675B & 17 & 109 & 48 & 125 \\
% \midrule
% Granite4-Small & 14 & 94 & 51 & 66 \\
% LLaMA-3.3-70B-Inst & 6 & 79 & 26 & 63 \\
% Qwen2.5-72B-Inst & 18 & 138 & 30 & 119 \\
% \bottomrule
% \end{tabular}
% \vspace{-3ex}
% \caption{
% Raw counts of extraction errors and hallucinations for samples in the Gnd error bucket in Table \ref{tab:sieve_success_combined}.
% }

% \label{tab:cap1-cap2-ragas-dialogue0-counts}
% \end{table}

\begin{figure*}[h]
    \centering
    \includegraphics[width=1.0\linewidth]{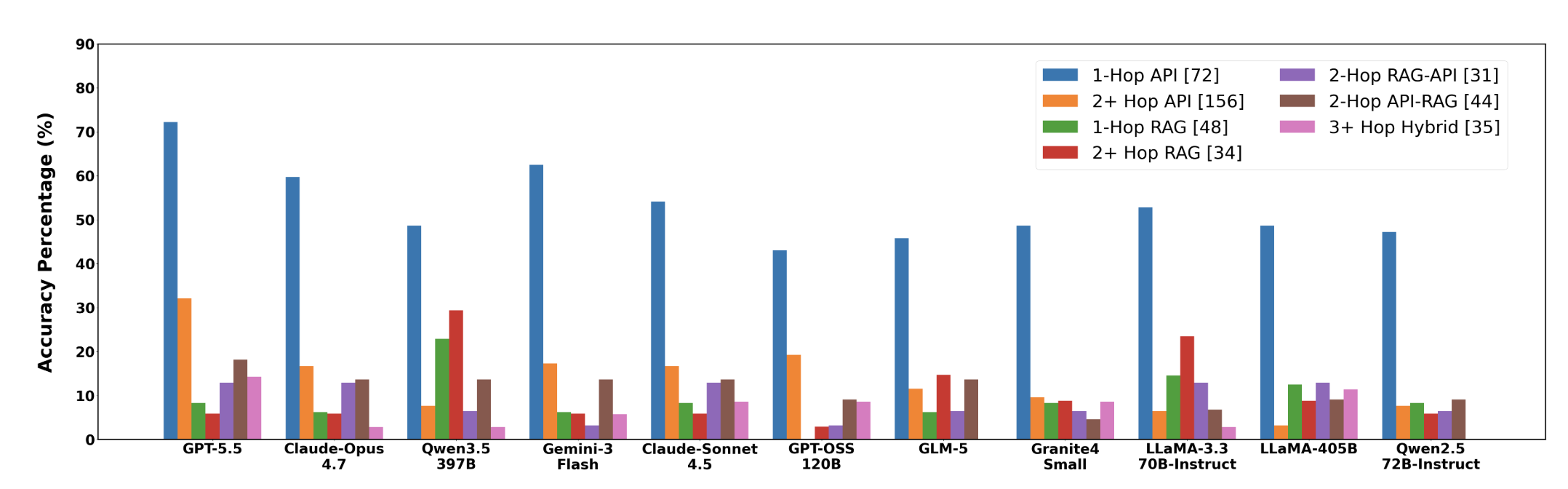}
    \caption{Model Accuracy Rates by Interaction Types}
    \label{fig:multiturn_by_question_type}
\end{figure*}
\subsection{Analysis}
\noindent{\bf API Styles:} Table \ref{tab:sieve_success_combined} shows the percentage of instances surviving each stage of tool calling across Dashboard APIs, SEL and SLOT APIs. As can be seen the errors in tool-calling for SEL and SLOT follow very different patterns. In SEL, once a correct tool is identified, it is usually successful at populating the argument names and their values as evidenced by the relatively flat error profile across {\em Tool} (tool identification), {\em ArgN} (argument identification) and {\em ArgV} (argument parameterization). On the other hand, errors can be traced to all stages of tool-calling as evidenced by the relatively steeper drop in the surviving instances at the ArgV stage.
%Though closed models have a much improved performance on all three API styles than compared to the results posted in \cite{elder2026liveapibench2500live} where GPT-5.5 is able to get 70\% accuracy as compared to the average accuracy of 50\% reported by \cite{elder2026liveapibench2500live} for Dashboard APIs.

Beyond tool-selection failures, grounding errors account for a large share of remaining mistakes. We categorize these into two types (Table~\ref{tab:cap1-cap2-ragas-dialogue0-counts}): (1)~extraction failures, where models inaccurately parse or truncate long tool responses, and (2)~hallucinations, where models generate information unsupported by the query or tool output. Hallucinations dominate grounding errors in both BI and
Dashboard APIs, though Dashboard APIs also exhibit frequent extraction failures likely due to longer structured responses that models tend to truncate.%Table\ref{tab:cap1-cap2-ragas-dialogue0-counts} further indicates that closed-source models demonstrate a stronger tendency to truncate tool responses. %Additionally, despite achieving the same successful tool-calling rate of 81.8 on Dashboard APIs, Gemini-3-Flash exhibits significantly higher extraction error and hallucination rates compared to GPT-5.5 (Table~4), resulting in an approximately 10-point gap in overall performance.

\noindent{\bf Multi-hop and Multi-Source Reasoning:}
Figure \ref{fig:multiturn_by_question_type} shows how the performance of models varies when questions require different depths of multi-hop reasoning. 1-hop APIs are the same task as evaluating the dashboard queries\footnote{Note we use a different set of queries in this collection} and performance drops when models encounter 2-hop API questions. Questions that require the use of the retriever have very low-performance though many models show a spike in performance on 2-hop RAG-API questions (eg: Gemini-3-flash-preview. This is likely explained by the relatively strong performance of some models on the dashboard APIs, and thus, once the correct intermediate answer is identified using the tool-call, the retrieval query is likely to be more successful. Interestingly, we find that on questions that require a single document retriever call (1-hop RAG), GPT-OSS-120B tries to directly return the answer from parameter knowledge, though when the question appears to require multiple hops, it answers the question. %We hypothesize that since the questions for 1-hop RAG are very Wikipedia-entity focussed the model skips the tool call (we don't see this problem on 1-hop API, where back-end database-specific entities/facts might be present more frequently in the question). 
%It is also interesting that the performance of Gemini-3-flash-preview shoots up on 2-hop API-RAG as compared to other hybrid hop-patterns. This is likely explained by the relatively strong performance of Gemini-3-flash-preview on the dashboard APIs (Tool Selection Capability), and thus, once the correct intermediate answer is identified using the tool-call, the retrieval query is likely to be more successful. 
Lastly, Qwen3.5-397B and Llama-3.3-70B-Instruct are best performing models for RAG-only hops, with GPT5.5 continuing to be the strongest model for multi-source hops. 

We also studied the performance on a subset where a mulit-hop reasoning chain ended with an API invocation. We find that on large models about 40-45\% of the errors can be tracked to answer-extraction/grounding though this increases to nearly 70-75\% when the hops involve RAG followed by an API. See Appendix \ref{app:additional_analysis} for details.

% \begin{enumerate}
%     \item Main Results
% \begin{itemize}
%      \item Business Intelligence - schema-bound
%      \item Business Intelligence - schema agnostic
%      \item Dashboard
% \end{itemize}
% \item Variation by number of tools available
% \item Domain-level analysis
% \item Error Analysis
% \end{enumerate}

\section{Conclusion}

\textcolor{black}{We presented \name{}, a benchmark for evaluating agentic reasoning across executable APIs and document collections spanning 62 domains. Unlike prior benchmarks that evaluate tool calling, retrieval, multi-hop reasoning, or policy adherence in isolation, \name{} is the first to require all of these within a single reasoning chain---agents must compose nested API sequences, ground information across structured and unstructured sources, and respect natural-language tool-use constraints, all verified by re-executing predictions against live database-backed APIs that accommodate multiple valid solution paths. By organizing tasks into three progressively challenging settings and evaluating models under a fixed ReAct harness, we isolate reasoning capabilities from agent architecture.}

\textcolor{black}{Our experiments yield three key findings. First, API interaction paradigm strongly influences difficulty: models that excel on endpoint-style APIs often struggle on compositional business-intelligence APIs, and vice versa, indicating that no single capability underlies tool-use proficiency. Second, multi-hop reasoning remains fragile---most models lose over 50\% accuracy as reasoning depth increases, with failures concentrating at language-mediated steps (entity disambiguation, cross-source grounding, schema alignment) rather than tool invocation mechanics. Third, tool-use policy adherence is a critical weakness: when policies render questions unanswerable, even frontier models fail to recognize this, with accuracy falling as low as 2.4\%.}

\textcolor{black}{These results suggest that improving agentic systems requires advances in compositional reasoning and constraint interpretation, not merely better tool-calling interfaces.}

% \section*{Limitations}

% \name{} evaluates models under a single agent paradigm (ReAct); alternative architectures (e.g., planning-based or tree-search agents) may yield different performance profiles. Our benchmark is constructed from BIRD-SQL domains and English-language documents, limiting generalization to other data modalities and languages. The tool-use policies we study are relatively simple natural-language constraints; real-world enterprise policies may involve more complex conditional logic. Finally, while our trajectory-level evaluation accommodates multiple valid paths, it relies on an LLM judge for borderline cases, introducing potential evaluation noise.

\bibliography{custom}

\appendix

\section{Overview}
In this appendix, we present additional error plots in Section \ref{app:additional_analysis}. Section \ref{app:data_generation} provides further details on data generation pipeline, data splits and data statistics. We provide a detailed human evaluation description in section \ref{app:human_eval} with the metrics used for the evaluation. Section \ref{app:eval_details} provides the LLM prompts used for evaluation and section \ref{app:model_parameters} provides details regarding the models evaluated. Section \ref{app:data_generation} and  \ref{app:multi_rag_pipeline} provided detailed description of the two synthetic data generation pipelines. Lastly, section \ref{app:benchmark_runner_and_harness} covers in detail the experimental runs including the benchmark runner and live execution environment setup.
 
\section{Additional Error Analysis}
\label{app:additional_analysis}
\begin{figure}[h]
\centering
\includegraphics[width=\columnwidth]{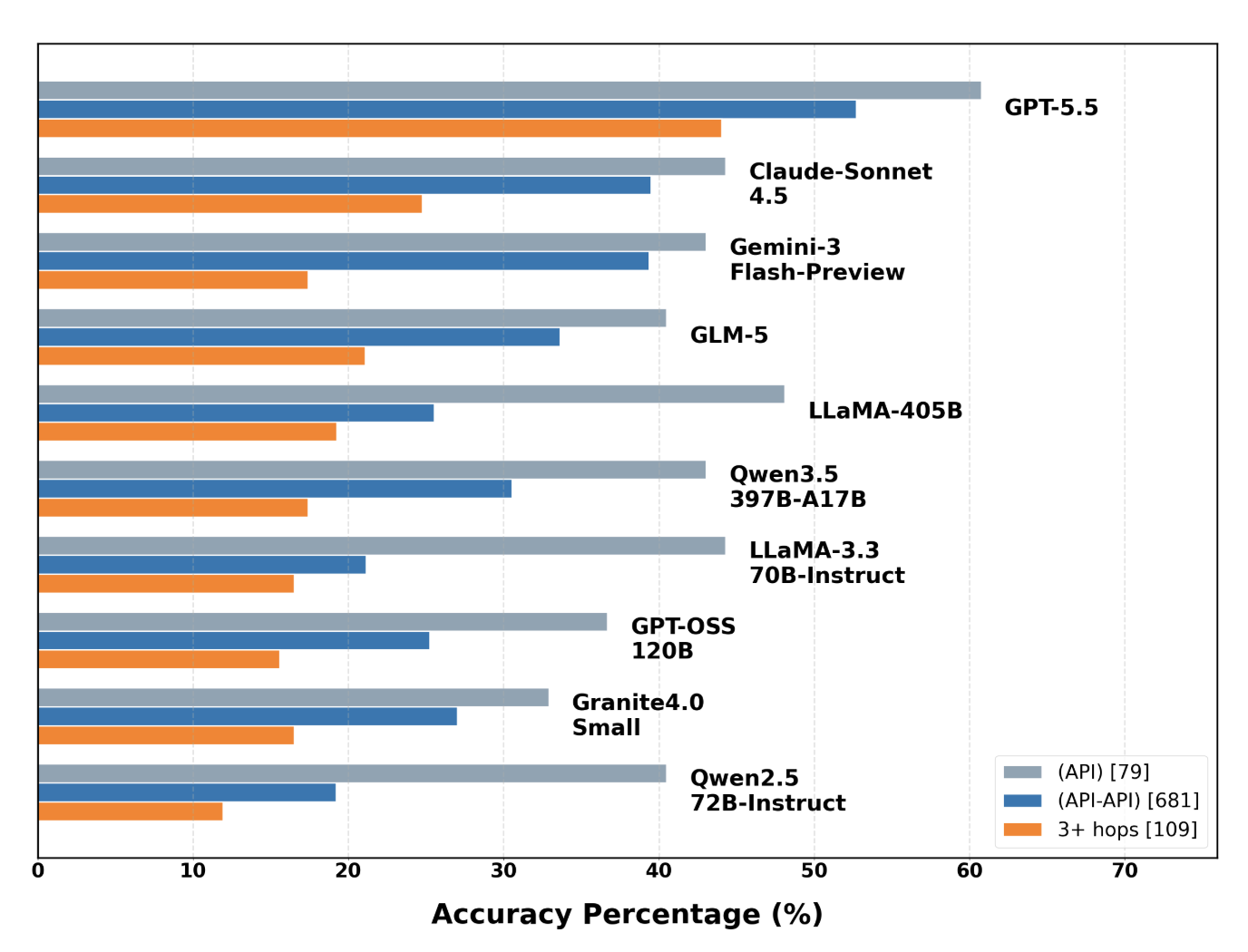}
\caption{Model Accuracy by number of hops for MultiHop Reasoning setting}
\label{fig:multihop_reaosning_error_chart}
\end{figure}

\paragraph{Performance for API hops} As can be seen from Figure \ref{fig:multihop_reaosning_error_chart} performance of the models decreases as number of hops increase.

\paragraph{Cross Source Entity Disambiguation} The nature of generation of the dataset (detailed in Section \ref{app:data_generation}) leads us to have hop-level labels. Thus, for every query we get the hop-level labels which are analyzed in Table \ref{tab:cross_grounding_performance}. For queries requiring extracting information from a document to fill the required entities for APIs is studied in Table \ref{tab:cross_grounding_performance}. Similarly, for queries needing only API hops, tool response from a previous hop is to be extracted to obtain the argument values of the succeeding hop. Table \ref{tab:entity_disambiguation} provides the accuracy of tool value extraction on such queries.

\begin{table}[t]
\centering
\scriptsize
\renewcommand{\arraystretch}{1.0}
\begin{tabular}{lrrr}
\hline
Model & Count & Total & \% \\
\hline
GPT-5.5 & 16 & 58 & 27.59 \\
Claude-Opus-4.7 & 17 & 58 & 29.31 \\
Gemini-3-Flash-Preview & 19 & 58 & 32.76 \\
Claude-Sonnet-4.5 & 13 & 58 & 22.41 \\
GLM-5.1 & 16 & 58 & 27.59 \\
\midrule
Qwen-3.5-397B & 17 & 58 & 29.31 \\
GPT-OSS-120B & 12 & 58 & 20.69 \\
LLaMA-405B & 12 & 58 & 20.69 \\
Mistral-Large-3-675B & 8 & 58 & 13.79 \\
\midrule
Granite-4h-small & 10 & 58 & 17.24 \\
LLaMA-3.3-70B-Instruct & 12 & 58 & 20.69 \\
Qwen-2.5-72B-Instruct & 12 & 58 & 20.69 \\
\hline
\end{tabular}

\caption{Cross source grounding performance comparison for Multisource queries in multihop multisource setting.}
\label{tab:cross_grounding_performance}
\end{table}

\begin{table}[t]
\centering
\scriptsize
\renewcommand{\arraystretch}{1.0}
\setlength{\tabcolsep}{2pt}
\begin{tabular}{l|cc|cc|cc}
\hline
\textbf{Model} &
\multicolumn{2}{c|}{\shortstack{\textbf{MultiHop}\\\textbf{Reasoning}}} &
\multicolumn{2}{c|}{\shortstack{\textbf{MultiHop MultiSource}\\\textbf{Reasoning}}} &
\multicolumn{2}{c}{\textbf{Combined}} \\
\cline{2-7}
& Cnt & \% & Cnt & \% & Cnt & \% \\
\hline
GPT-5.5                  & 473 & 59.87 & 125 & 58.41 & 598 & 59.56 \\
Claude-Opus-4.7          & 441 & 55.82 & 119 & 55.61 & 560 & 55.78 \\
Gemini-3-Flash-Preview   & 426 & 53.92 & 101 & 47.20 & 527 & 52.49 \\
Claude-Sonnet-4.5        & 432 & 54.68 &  97 & 45.33 & 529 & 52.69 \\
GLM-5.1                  & 385 & 48.73 &  95 & 44.39 & 480 & 47.81 \\
\midrule
Qwen-3.5-397B            & 379 & 47.97 & 102 & 47.66 & 481 & 47.91 \\
GPT-OSS-120B             & 230 & 29.11 &  62 & 28.97 & 292 & 29.08 \\
LLaMA-405B               & 234 & 29.62 &  20 &  9.35 & 254 & 25.30 \\
Mistral-Large-3-67B      & 235 & 29.75 &  47 & 21.96 & 282 & 28.09 \\
\midrule
Granite-4h-small         & 234 & 29.62 &  30 & 14.02 & 264 & 26.29 \\
LLaMA-3.3-70B-Instruct   &  61 &  7.72 &  31 & 14.49 &  92 &  9.16 \\
Qwen-2.5-72B-Instruct    & 194 & 24.56 &  50 & 23.36 & 244 & 24.30 \\
\hline
\end{tabular}
\caption{Entity disambiguation performance comparison across single-source
queries belonging to the multihop setting and multisource multihop setting,
as well as the combined evaluation set. The total numbers of evaluated queries
are 790, 214, and 1004 for MultiHop Reasoning, MultiHop MultiSource Reasoning,
and Combined, respectively.}
\label{tab:entity_disambiguation}
\end{table}

\paragraph{Predicted Tool Calls per model} Table \ref{tab:avg-predicted-tool-calls-combined} provides the number of tool calls per model per setting. Notably, the LLaMA model has fewer than 2 predicted calls per sample per setting, specifically having lesser than 2 predicted calls even for the multihop queries. On the contrary, models with better performance (i.e., GPT-5.5, GLM-5.1 were able to identify the need for multiple predicted tool calls for multihop and multihp multisource settings).

\begin{table}[t]
\centering
\scriptsize
\setlength{\tabcolsep}{2.5pt}
\renewcommand{\arraystretch}{1.0}

\begin{tabular}{
p{2.1cm}
c c c c c c
}
\toprule

\textbf{Model} &
\shortstack{\textbf{BI}\\\textbf{APIs}} &
\shortstack{\textbf{Dashboard}\\\textbf{APIs}} &
\shortstack{\textbf{Multi}\\\textbf{Hop}} &
\shortstack{\textbf{MultiHop}\\\textbf{MultiSource}} &
\shortstack{\textbf{Overall}\\\textbf{Avg.}} \\

\midrule

GLM-5.1                & 3.87 & 2.50 & 4.65 & 4.86 & 3.70 \\
GPT-5.5                & 3.66 & 2.06 & 5.19 & 5.76 & 3.69 \\
Claude-Sonnet-4.5      & 3.03 & 1.56 & 4.27 & 4.93 & 3.03 \\
Granite-4h-small       & 3.33 & 1.45 & 1.97 & 1.80 & 2.31 \\
Gemini-3-Flash         & 3.06 & 2.77 & 8.15 & 4.68 & 4.05 \\
GPT-OSS-120B           & 3.79 & 1.21 & 1.90 & 3.09 & 2.57 \\
LLaMA-3.3-70B          & 2.73 & 0.98 & 1.13 & 1.18 & 1.70 \\
LLaMA-405B             & 2.50 & 1.00 & 1.85 & 0.98 & 1.72 \\
% Mistral-Large3-675B   & 2.38 & 1.08 & 2.24 & 1.89 & 1.88 \\
Qwen-2.5-72B           & 2.98 & 1.40 & 3.42 & 2.38 & 2.48\\
Qwen-3.5-397B          & 6.07 & 3.12 & 7.87 & 6.64 & 5.53\\

\bottomrule
\end{tabular}

\vspace{-2ex}
\caption{
Average number of predicted tool calls per sample across benchmark settings and overall model-level average.
}
\label{tab:avg-predicted-tool-calls-combined}
\end{table}

% \begin{figure*}
% \centering
% \includegraphics[scale=0.16]{figures/CAP1_SEL_SLOT.png}
% \caption{Error patterns plot for SEL and SLOT Business Intelligence APIs. Plot for values presented in Table \ref{tab:sieve_success_combined}}
% \label{fig:task1_error_plot}
% \end{figure*}

% \begin{figure*}
% \centering
% \includegraphics[scale=0.2]{figures/CAP1_error_plot.png}
% \caption{Sieve of success plot for Business Intelligence API. Plot for values presented in Table \ref{tab:sieve_success_combined}}
% \label{fig:task1_error_plot}
% \end{figure*}

% \begin{figure*}
% \centering
% \includegraphics[scale=0.2]{figures/CAP2_error_plot.png}
% \caption{Sieve of success plot for Dashboard APIs. Plot for values presented in Table \ref{tab:sieve_success_combined}}
% \label{fig:task2_error_plot}
% \end{figure*}

\section{Data Generation}
\label{app:data_generation}
This section includes the details of the data generation pipeline and additional data statistics for this benchmark and the intended use of the dataset. Section \ref{app:data_sample} includes a sample of the dataset for every task. Section \ref{app:data_statistics} and section \ref{app:dataset_splits} provide details of the data splits within the tasks covered in this benchmark as well as a few more data statistics. Section \ref{app:query_generation} and Section \ref{app:multi_rag_pipeline} walk through the experimental details of the synthetic query generation for all the question types covered in this work.

\subsection{Intended Use}
We carefully reviewed all datasets and artifacts used in this work to ensure they do not contain personally identifiable information (PII), offensive content, or sensitive user data. The benchmark is constructed from publicly available research datasets, structured databases, and knowledge sources intended for academic use, and does not include real user conversations, private enterprise logs, or human subject data. Our use of existing resources, including BIRD-SQL, Wikidata5M, and related retrieval corpora, is consistent with their intended research and evaluation purposes. The released benchmark artifacts are intended solely for non-commercial research use in evaluating tool-using language agents and should not be deployed for production decision-making or surveillance applications. In accordance with these constraints, the dataset and accompanying benchmark framework would be released under the CC-BY-NC-SA-4.0 license.

\subsection{Additional Data Statistics}
\label{app:data_statistics}
\begin{figure*}[t]
  \centering
  \includegraphics[width=1.0\linewidth]{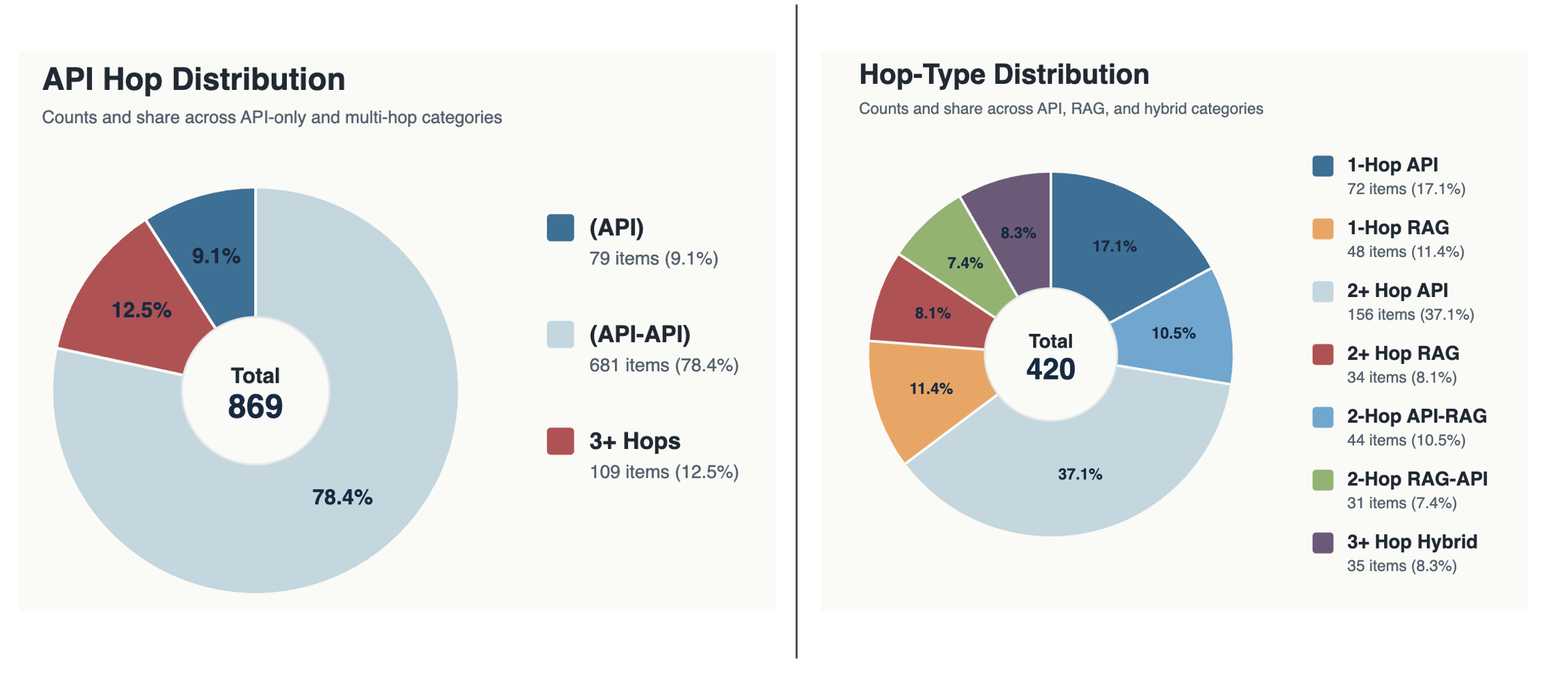}
  \caption{Question Type Distribution for Multihop setting and the MultiHop Multisource questions}
  \label{fig:question-type-distribution}
\end{figure*}

As shown in Figure \ref{fig:question-type-distribution} the 869 questions in multihop setting have $2$-$5$ API hops included in its dataset. Also, of the 664 samples in the Multihop multisource with policy adherence setting dataset 420 samples donot have any policy applied to them. The distribution of question types for these samples is also provided in Figure \ref{fig:question-type-distribution}.

\begin{table}[t]
\centering
\caption{Benchmark domains grouped by semantic clusters.}
\label{tab:domain-clusters}
\scriptsize
\setlength{\tabcolsep}{2pt}
\renewcommand{\arraystretch}{1.0}
\begin{tabular}{p{2.7cm} p{4.7cm}}
\toprule
\textbf{Cluster} & \textbf{Domains} \\
\midrule

Education \& Students &
california\_schools, student\_loan, university, computer\_student, college\_completion \\

Movies/TV \& Social Media &
movie, movie\_platform, movie\_3, movies\_4, simpson\_episodes, disney, law\_episode, movielens, talkingdata, music\_tracker, music\_platform\_2, card\_games, video\_games, superhero \\

Sports \& Athletics &
formula\_1, professional\_basketball, european\_football\_1, european\_football\_2, olympics, ice\_hockey\_draft, hockey, soccer\_2016 \\

Retail \& Commerce &
book\_publishing\_company, sales\_in\_weather \\

Food \& Beverage &
restaurant, food\_inspection, cookbook, beer\_factory, craftbeer, menu \\

Technology \& Software &
app\_store, codebase\_comments, image\_and\_language \\

Health \& Medicine &
toxicology, mental\_health\_survey, thrombosis\_prediction, genes \\

Geography \& Demographics &
mondial\_geo, world, address, world\_development\_indicators \\

Transportation \& Mobility &
airline, trains, bike\_share\_1, cars \\

Finance \& Economics &
financial, coinmarketcap, debit\_card\_specializing \\

Government \& Public Services &
legislator, chicago\_crime, public\_review\_platform \\

Literature \& Publishing &
books, authors, shakespeare, language\_corpus, citeseer \\

\bottomrule
\end{tabular}
\end{table}

\subsection{Benchmark Data split description}
\label{app:dataset_splits}
The BIRD-SQL dataset \cite{bird} contains over 12,751 unique question-SQL pairs. These queries are the base queries utilized for all the tasks in our dataset. Following is the logic used for query selection for the three settings in our benchmark
\begin{enumerate}
    \item Only a limited number of queries had entities which could have linkages with base Wikidata5m passages ~\citep{wang2021kepler} required to form a knowledge graph needed for Multihop and Multisource query generation. So, any query which could be connected via an retriever question was only used for the purpose of constructing API-RAG style joint reasoning queries. A detailed description of the knowledge graph construction and query generation is included in section\ref{app:query_generation}. These API-RAG queries occur only in the multihop multisource setting.
    \item Queries which could form (API-API) linkages based on the query entities and answer entities were reserved for multihop setting of the dataset.
    \item All the other queries which couldn't be added to the knowledge graphs due to entities which couldn't be linked to other queries or their answers were used as the queries for the Dashboard APIs task.
    \item As the nesting behaviour of Business Intelligence APIs (i.e. SEL and SLOT) already leads to a relatively more difficult task the queries used for this split of the dataset could be overlapping with other settings.
    \item No component BIRD-SQL are shared between the tuning and test set of our dataset.
\end{enumerate}

\subsection{Data Sample}
\label{app:data_sample}
\begin{figure*}[t]
\centering
\includegraphics[width=\textwidth]{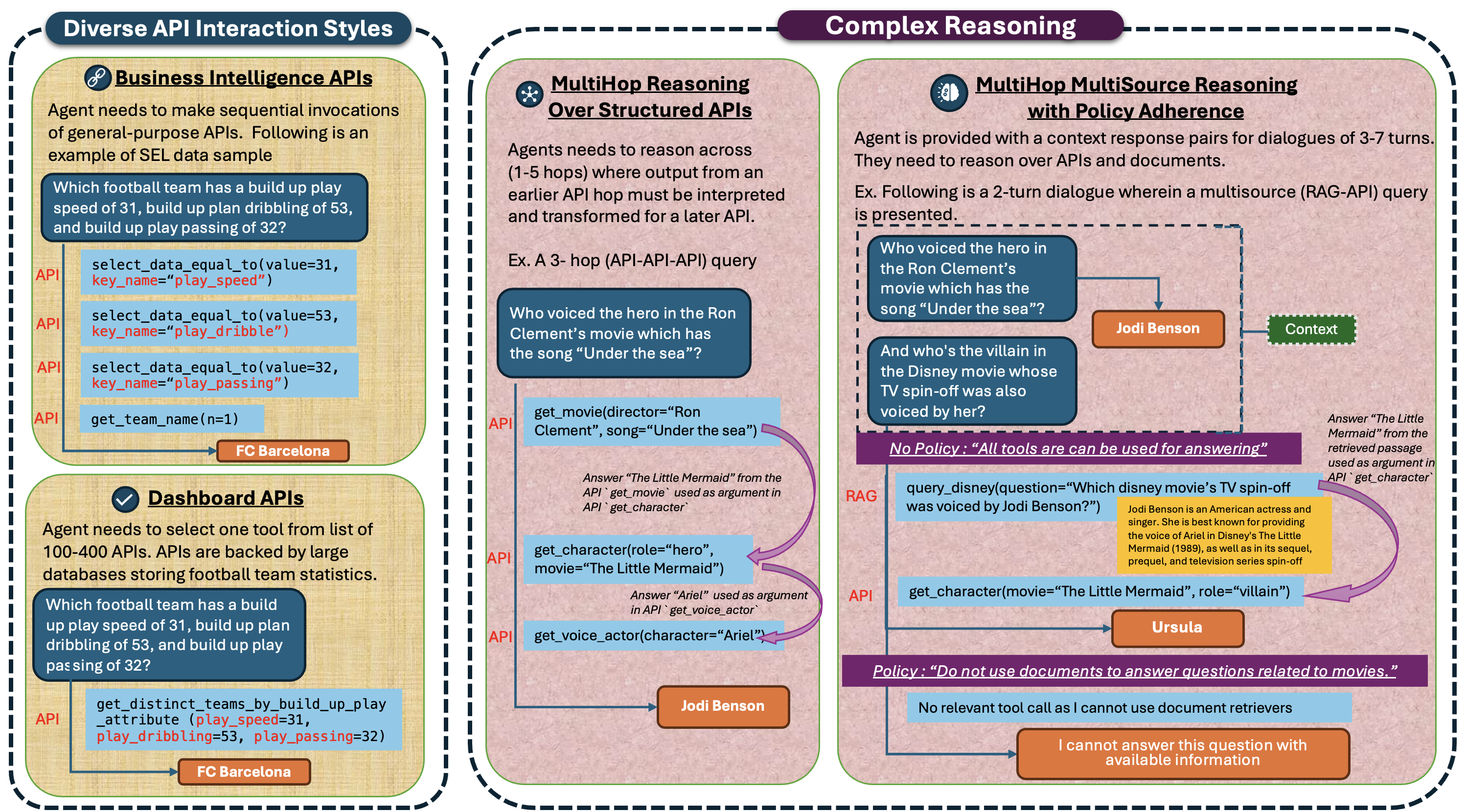}
\caption{Data sample demonstrating a sample for each of the settings (i) SEL Business Intelligence API type (ii) Dashboard API type (iii) MultiHop Reasoning (iv) MultiHop MultiSource Reasoning with Policy Adherence}
\label{fig:data_sample}
\end{figure*}

Figure \ref{fig:data_sample} shows a data sample for SEL Business Intelligence API and the ground truth tool calls required to answer the query as well as a data sample from the Dashboard APIs collection and structure of APIs required to answer queries. It also shows a data sample for a multihop reasoning query from the Disney domain in our dataset with the 2-hop reasoning chain required to answer the query. Lastly, it also includes a (RAG-API) reasoning query from the MultiHop MultiSource reasoning setting. The effect of policy on the final answer is demonstrated in the data sample, wherein the policy "Do not use documents to answer questions related to movies" renders the ground truth answer for the query to turn to "No relevant tool call to answer the given query." from it's original answer.

\begin{figure}[t]
  \centering
  \includegraphics[width=0.75\linewidth]{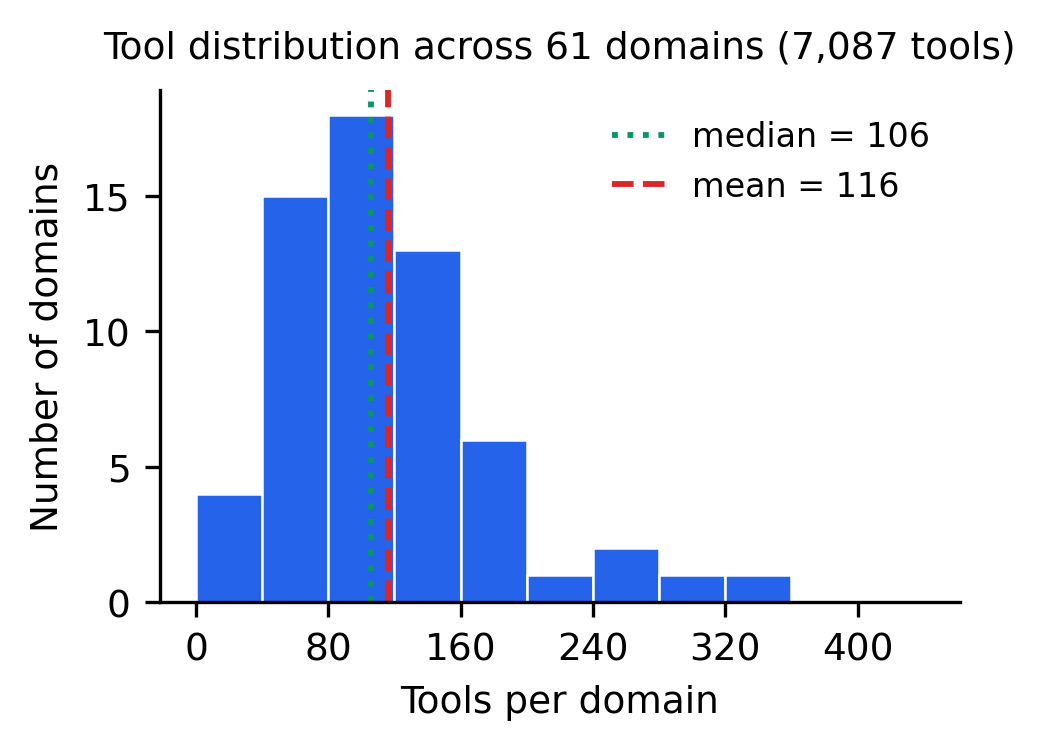}
  \caption{Distribution of the number of tools per domain across the
  $62$ benchmark domains ($7{,}087$ tools total). The distribution is
  right-skewed (mean $116$, median $106$), with most domains exposing
  $40$--$160$ tools and a thin tail of tool-rich domains.}
  \label{fig:tool-hist}
\end{figure}

\subsection{Dashboard APIs Distribution Across Domains}
\label{app:tool-distribution}

Our benchmark spans $62$ domains, each exposing a set of REST API tools that an agent may invoke. In total the benchmark comprises $7{,}087$ tools, with a mean of $116.2$ tools per domain
(median $106$, standard deviation $63.1$). As shown in Figure \ref{fig:tool-hist} the distribution is
right-skewed: the interquartile range is $[74, 154]$ tools, while a
small number of tool-rich domains form a long upper tail. The smallest
domain, \texttt{craftbeer}, exposes only $6$ tools, whereas the largest,
\texttt{public\_review\_platform}, exposes $328$---a $55\times$ spread.
This heterogeneity is intentional: it lets us evaluate agent behaviour
both in compact action spaces and in large ones where tool retrieval
and selection become non-trivial.

% ----------------------- Histogram figure --------------------------
\begin{table*}[t]
  \centering
  \caption{Number of tools per domain for all $62$ benchmark domains,
  sorted in descending order of tool count.}
  \label{tab:tool-counts}
  \footnotesize
  \setlength{\tabcolsep}{4pt}
  \resizebox{\textwidth}{!}{%
  \begin{tabular}{lr@{\hskip 1.5em}lr@{\hskip 1.5em}lr}
    \toprule
    Domain & \#Tools & Domain & \#Tools & Domain & \#Tools \\
    \midrule
    public\_review\_platform & 328 & beer\_factory & 124 & ice\_hockey\_draft & 79 \\
    mondial\_geo & 283 & olympics & 122 & sales\_in\_weather & 76 \\
    movie\_3 & 270 & european\_football\_2 & 121 & cars & 75 \\
    soccer\_2016 & 244 & toxicology & 117 & college\_completion & 75 \\
    hockey & 206 & image\_and\_language & 116 & book\_publishing\_company & 72 \\
    simpson\_episodes & 190 & university & 115 & computer\_student & 66 \\
    card\_games & 182 & language\_corpus & 115 & app\_store & 63 \\
    talkingdata & 179 & bike\_share\_1 & 110 & music\_platform\_2 & 60 \\
    chicago\_crime & 174 & superhero & 109 & debit\_card\_specializing & 60 \\
    legislator & 170 & codebase\_comments & 106 & cookbook & 59 \\
    student\_loan & 167 & law\_episode & 106 & european\_football\_1 & 56 \\
    thrombosis\_prediction & 159 & restaurant & 105 & movie & 46 \\
    movie\_platform & 156 & disney & 104 & coinmarketcap & 46 \\
    professional\_basketball & 156 & financial & 102 & music\_tracker & 45 \\
    books & 155 & menu & 100 & mental\_health\_survey & 45 \\
    formula\_1 & 152 & movielens & 96 & trains & 38 \\
    world\_development\_indicators & 147 & shakespeare & 94 & genes & 23 \\
    video\_games & 143 & world & 89 & citeseer & 19 \\
    address & 139 & california\_schools & 89 & bpo & 13 \\
    movies\_4 & 138 & airline & 85 & craftbeer & 6 \\
    authors & 133 & food\_inspection & 82 & & \\
    \bottomrule
  \end{tabular}%
  }
\end{table*}

\section{Human Evaluation Details}
\label{app:human_eval}

We perform detailed human evaluation for generated \textbf{Multi-hop Reasoning} and \textbf{Multi-hop Multi-Source Reasoning} questions to validate the compositional quality of automatically generated reasoning chains. Each question is independently annotated by $3$ annotators. Annotators are shown:
(i)~the generated merged query,
(ii)~the component queries used during composition,
(iii)~ground-truth tool chains,
(iv)~tool responses,
(v)~retrieved documents for RAG-based settings, and
(vi)~the final ground-truth answer.

The annotation interface supports rubric-guided scoring with hoverable metric definitions and keyboard-assisted annotation to improve consistency and annotation throughput.

\paragraph{Evaluation Metrics.}
Our evaluation dimensions are motivated by prior work on compositional reasoning~\citep{trivedi2022musique,yang2018hotpotqa}, query generation~\citep{agentcq2024,llmenhancedquery2025,qaexpand2025}, and faithfulness evaluation in RAG systems~\citep{es2024ragas}.

\begin{itemize}
    \item \textbf{Faithfulness Score:}
    Measures whether the generated query contains only information grounded in the component queries without introducing unsupported entities, relations, or constraints.

    \item \textbf{Logical Consistency Score:}
    Measures whether the generated query contains contradictory or logically incompatible constraints while preserving a coherent reasoning chain.

    \item \textbf{Answer Leakage Score:}
    Measures whether the generated query explicitly or implicitly reveals intermediate-hop or final answers, making the reasoning process trivial.

    \item \textbf{Context Sufficiency Score:}
    Measures whether the generated question contains sufficient information to execute the required reasoning chain and populate tool arguments.

    \item \textbf{Context Sufficiency Score (Applicable only for questions with documents):}
    Measures whether retrieved documents contain sufficient evidence to answer the retrieval-grounded component of the query.

    \item \textbf{Cross-Hop Entity Consistency Score (Applicable only for questions with documents):}
    Measures whether entities propagated between API and retrieval hops are correctly inferred and grounded in retrieved evidence.
\end{itemize}

\paragraph{Scoring Scale}
All metrics are evaluated using a rubric-based ordinal $1$--$4$ scoring scheme. The scoring framework is inspired by graded factuality and coherence evaluation used in summarization and RAG evaluation~\citep{maynez2020faithfulness}. Lower scores correspond to severe reasoning failures, while higher scores correspond to fully coherent and grounded compositional reasoning.

Table \ref{tab:human_eval_agreement} provides granular analysis of scores provided by annotators. We report percentage agreement due to ordinal multi-annotator scoring.

\begin{table}[t]
\centering
\footnotesize
\setlength{\tabcolsep}{3pt}
\begin{tabular}{lccc}
\toprule
\textbf{Metric} & \textbf{Agr.} & \textbf{F1} & $\mathbf{\kappa}$ \\
\midrule
\multicolumn{4}{c}{\textit{Multi-hop Reasoning}} \\
\midrule
Faithfulness & 71.7 & 0.640 & -0.088 \\
Logical Consistency & 71.7 & 0.695 & 0.014 \\
Answer Leakage & 88.3 & 0.888 & 0.331 \\
Context Sufficiency & 75.0 & 0.718 & 0.307 \\
\midrule
\multicolumn{4}{c}{\textit{Multi-hop Multi-source Reasoning}} \\
\midrule
Faithfulness & 95.0 & 0.958 & -0.023 \\
Logical Consistency & 80.0 & 0.741 & -0.039 \\
Answer Leakage & 91.7 & 0.941 & -0.031 \\
Context Sufficiency & 88.3 & 0.891 & -0.051 \\
Retrieval Sufficiency & 93.3 & 0.926 & 0.879 \\
Cross-Hop Entity Consistency & 91.7 & 0.917 & 0.868 \\
\bottomrule
\end{tabular}
\caption{Inter-annotator agreement for MultiHop; and MultiHop MultiSource human evaluation task. Agreement denotes exact agreement (\%).}
\label{tab:human_eval_agreement}
\end{table}

\paragraph{Metric provided to annotators}

Metrics for the multihop reasoning task
\begin{lstlisting}
METRICS = {
    "faithfulness": {
        "title": "Faithfulness Score",
        "desc": "Does the merged query contain only information grounded in the component queries, without introducing unsupported facts?",
        "scale": {
            1: "Completely hallucinated: major unsupported entities or relations introduced",
            2: "Harmful hallucination: incorrect info that breaks reasoning",
            3: "Minor hallucination: small issues but mostly correct",
            4: "No hallucination: fully grounded in component queries"
        }
    },
    "logical_consistency": {
        "title": "Logical Consistency",
        "desc": "Is the merged query logically consistent, with no contradictions between its components, and aligned with the reasoning chain implied by intermediate queries?",
        "scale": {
            1: "Completely inconsistent and contradictory leading it to be unanswerable",
            2: "Major inconsistency to the level of being misleading",
            3: "Minor inconsistency not affecting answerability",
            4: "Fully consistent"
        }
    },
    "answer_leakage": {
        "title": "Answer Leakage",
        "desc": "Does the merged query explicitly or implicitly reveal the answer entity, making the reasoning task trivial? (i.e., Is the answer entity accidentally mentioned in the query, or does the merged query leak any answer to the hop question?)",
        "scale": {
            1: "Complete leakage (answer directly present)",
            2: "Strong leakage (very obvious answer)",
            3: "Minor hints present",
            4: "No leakage"
        }
    },
    "context_sufficiency": {
        "title": "Context Sufficiency",
        "desc": "Does the merged query contain sufficient context and constraints to be answerable via the intended tool or retrieval pipeline? (i.e., Does the merged question have enough context present to be answered as well as does the query have enough context to populate the arguments of the tools?)",
        "scale": {
            1: "Not answerable",
            2: "Major missing context",
            3: "Mostly sufficient",
            4: "Fully sufficient"
        }
    }
}
\end{lstlisting}

Metrics for multihop multisource reasoning task
\begin{lstlisting}
METRICS = {
    "faithfulness": {
        "title": "Faithfulness Score",
        "desc": "Does the merged query contain only information grounded in the component queries, without introducing unsupported facts?",
        "scale": {
            1: "Completely hallucinated: major unsupported entities or relations",
            2: "Harmful hallucination: incorrect info that breaks reasoning",
            3: "Minor hallucination: small issues but mostly correct",
            4: "No hallucination: fully grounded in component queries"
        }
    },
    "logical_consistency": {
        "title": "Logical Consistency",
        "desc": (
        "Does the merged query avoid logically incompatible conditions or contradictions? "
        "The different parts of the query should be simultaneously satisfiable and should "
        "form a valid reasoning chain. For example, a question like "
        "'Name a city on Earth which lies above the equator and is in Australia?' "
        "contains logically incompatible constraints."
    ),
        "scale": {
            1: "Completely inconsistent: logically incompatible component queries or conditions that cannot be satisfied together were merged",
            2: "Major inconsistency: the query contains impossible or directly contradictory conditions that make the question invalid or unanswerable",
            3: "Minor inconsistency: the query is mostly logically valid but contains small ambiguities, weak conflicts, or mildly confusing constraints",
            4: "Fully consistent"
        }
    },
    "answer_leakage": {
        "title": "Answer Leakage",
        "desc": "Does the merged query explicitly or implicitly reveal the answer entity, making the reasoning task trivial? (i.e., Is the answer entity accidentally mentioned in the query, or does the merged query leak any answer to the hop question?)",
        "scale": {
            1: "Complete leakage (answer directly present)",
            2: "Strong leakage (very obvious answer)",
            3: "Minor hints present",
            4: "No leakage"
        }
    },
    "context_sufficiency": {
        "title": "Context Sufficiency",
        "desc": "Does the merged query contain sufficient context and constraints to be answerable via the intended tool or retrieval pipeline? (i.e., Does the merged question have enough context present to be answered as well as does the query have enough context to populate the arguments of the tools?)",
        "scale": {
            1: "Not answerable: entities missing in the merged query or insufficient context to answer the question",
            2: "Major missing context: all entities present but insufficient context significantly hindering answerability",
            3: "Mostly sufficient: minor missing context that does not significantly hinder answerability",
            4: "Fully sufficient"
        }
    },
    "retrieval_sufficiency": {
        "title": "Retrieval Sufficiency Score",
        "desc": "Do the ground truth documents have sufficient information to answer the RAG query? (Mark '0' if no RAG component in query.)",
        "scale": {
            0: "Not applicable (e.g., no retrieval needed for this query)",            
            1: "GT document have no relevant information",
            2: "GT document have some missing information",
            3: "GT document have some missing information which is common sense knowledge",
            4: "No information missing"
        },
    },
    "cross_hop_entity_consistency": {
        "title": "Cross-Hop Entity Consistency Score",
        "desc": "Are entities required by the arguments of the succeeding or preceding API tool calls correctly inferred and grounded in the retrieved documents or retriever questions? (Mark '0' if no RAG component in query.)",
        "scale": {
            0: "Not applicable (e.g., no retrieval needed for this query)",
            1: "Not answerable",
            2: "Majorly missing context / entities cannot be answered without these entities",
            3: "Mostly sufficient have some missing information which is common sense knowledge",
            4: "Fully sufficient",
        },
    },
}
\end{lstlisting}

\section{Evaluation Details}
\label{app:eval_details}

\paragraph{Policy Adherence Check}
For policy-constrained tasks, we inspect the agent's tool-call trace and verify that no disallowed sources were consulted. This is deterministic: each tool call is tagged with its source type (API or retrieval collection due to the nature of our generation pipeline), and we check against the per-query policy specification. An agent may produce the correct final answer while violating a policy -- such traces are marked as failures.

\paragraph{LLM Judge Prompts} The RAGAS \cite{es2024ragas} answer correctness prompt is used for factual correctness in stage 2. Following is the groundedness prompt used for stage 2 :

\begin{lstlisting}
GroundednessPrompt ="""
The following tasks each contains document and a response. The response is supposed to rely on the document for its source of information, optionally using common sense knowledge and common sense inference, but it may fail this, and instead contain substantial claims that are not grounded in the document or common sense knowledge.

Your task is to assess whether the response is entirely grounded in the document, grounded in the document plus common sense knowledge and reasoning, or ungrounded. To make this determination, perform the following steps:
1. Identify all substantial claims in the response:
   - Ignore non-substantial claims, such as greetings or self-descriptions such as "I'm a helpful assistant",
   - Try to formulate each claim in a stand-alone form with all pronouns and other references resolved;
2. Assess the grounding of each of these claims:
   - If it is essentially a rephrasing of information from the document, or can be derived from such information by trivial common-sense reasoning, it is grounded,  This is so even if it contradicts other parts of the document.  
   - If it relies on, in additional to information from the document, additional non-trivial common sense knowledge or common sense reasoning, it is partially grounded,
   - If a claim is about the provided document, or about the agent\'s state of knowledge, with the effect of not being able to answer the user inquiry, it is grounded if and only if the required information is indeed lacking in the document.
   - If a claim cannot be derived directly from the document or indirectly with help of common sense knowledge and reasoning, it is ungrounded;
3. Make the overall decision according to:
   - If at least one claim is not grounded, the response is not grounded (Note that this is not a case of partially grounded);
   - Otherwise if at least one claim is partially grounded, the response is partially grounded;
   - Otherwise the response is grounded.

Pay attention that: Even if the document contains the keyword of response, it does not mean the response is grounded, and you have to make decision based on 1,2,3 above.

Your final conclusion should be written in two lines:
- The first line contains one of the following labels  [yes, partial, no, unsure],
  - "yes" is for grounded,
  - "partial" is for grounded with non-trivial common sense knowledge or reasoning,
  - "no" is for ungrounded,
  - "unsure" is for the situations where the document, conversation or response contain ambiguities such that different interpretations lead to different conclusions about groundedness;
- The second line contains an explanation of your answer as short as possible.
  
Here is the document starting with <doc> and end with </doc>
<doc> 
{doc}
</doc> 

Here is the response starting with <response> and end with </response>.
<response>
{response}
</response>

Now write your final conclusion following below format:
<conclusion>
choose a label from [yes, partial, no, unsure] based on your analysis of given document and response.
- "yes" is for grounded,
- "partial" is for grounded with non-trivial common sense knowledge or reasoning,
- "no" is for ungrounded,
- "unsure" is for the situations where the document, conversation or response contain ambiguities such that different interpretations lead to different conclusions about groundedness;
</conclusion>
"""
\end{lstlisting}

\section{Model Parameters}
\label{app:model_parameters}

Table \ref{tab:model-compute} provides details regarding the models evaluated in our experiments with their checkpoint identifiers, and serving infrastructure.\\ Majority of the open-weight models are served via a vLLM interface hosted on NVIDIA GPUs. The closed models are access via various providers.

\begin{table}[t]
\centering
\scriptsize
\setlength{\tabcolsep}{3pt}
\renewcommand{\arraystretch}{1.02}

\begin{tabular}{
p{1.9cm}
p{0.9cm}
p{2.8cm}
p{1.8cm}
}
\toprule
\textbf{Model} & \textbf{\#Params} & \textbf{Model Checkpoint} & \textbf{Infrastructure} \\
\midrule

GPT-5.5 &
Closed Model &
-- &
Azure OpenAI \\

Claude-Opus-4.7 &
Closed Model &
-- &
AWS Bedrock \\

Gemini3-Flash &
Closed Model &
--  &
GCP Vertex AI \\

Claude-Sonnet-4.5 &
Closed Model &
--  &
AWS Bedrock \\

GLM-5.1 &
Closed Model &
-- &
AWS \\

\midrule

Qwen3.5-397B &
397B &
Qwen/Qwen3.5-397B-A17B-FP8 &
vLLM + NVIDIA GPUs \\

LLaMA-405B &
405B &
meta-llama/llama-3-1-405b-instruct-fp8 &
vLLM + NVIDIA GPUs \\

GPT-OSS-120B &
120B &
openai/gpt-oss-120b &
vLLM + NVIDIA GPUs \\

% Mistral-Large3-675B &
% 675B &
% mistralai/Mistral-Large-3-675B-Instruct-2512-NVFP4 &
% vLLM + NVIDIA GPUs \\

\midrule

Granite-4h-Small &
32B &
ibm-granite/granite-4.0-h-small &
vLLM + NVIDIA GPUs \\

LLaMA-3.3-70B-Instrust &
70B &
meta-llama/llama-3-3-70b-instruct &
vLLM + NVIDIA GPUs \\

Qwen2.5-72B-Instrust &
72B &
Qwen/Qwen2.5-72B-Instruct &
vLLM + NVIDIA GPUs \\

\bottomrule
\end{tabular}

\vspace{-2ex}
\caption{
Models evaluated in our experiments, along with parameter counts, checkpoint identifiers, and serving infrastructure.
Closed-source models were accessed through commercial APIs,
while open-weight models were served using vLLM on NVIDIA GPUs.
}

\label{tab:model-compute}
\end{table}

\input{appendix/agent_architecture}

\input{appendix/multiturn_multihop}

\input{appendix/multiturn_rag}

\end{document}

%% file: appendix/agent_architecture.tex
\section{Benchmark Runner and Agent Implementation}
\label{app:benchmark_runner_and_harness}
\paragraph{Benchmark Runner.}
The runner (\texttt{benchmark\_runner.py}) orchestrates the full evaluation
lifecycle. Given one or more capability IDs, it loads the benchmark items,
groups them by domain, and processes each domain against a freshly connected
MCP server, with connection settings read from a YAML config. For each domain it
opens an MCP session over stdio, retrieves the tool list, and verifies it
against the committed checksum for the \texttt{(capability, domain)} pair---a
mismatch aborts the run before any query executes. The verified tools are then
wrapped into typed LangChain tools and handed to the agent.

\begin{lstlisting}[caption={Per-domain setup: connect, verify integrity, wrap tools.}]
async with AsyncExitStack() as stack:
    session = await stack.enter_async_context(
        create_client_and_connect(cfg, domain)
    )

    # Client-side integrity check before any query runs.
    raw_tools = (await session.list_tools()).tools
    verify_checksum(capability_id, domain, raw_tools)

    tools = await MCPToolWrapper(session).get_tools()
    agent = _get_agent(capability_id, llm, tools)
\end{lstlisting}

Multi-hop settings additionally support \emph{tool-universe switching}: before
each query the runner invokes a \texttt{get\_data} tool with the item's UUID,
which loads the per-item dataset server-side and returns a handle injected into
the agent's state. Each query then runs under a fixed wall-clock timeout, with
timeouts and exceptions captured as error results so a single failure never
aborts the run.

\begin{lstlisting}[caption={Per-item dataset switching and timed agent run.}]
# Multi-hop: load this item's dataset server-side, inject the handle.
if get_data_tool:
    data = json.loads(
        await get_data_tool.ainvoke({"tool_universe_id": item.uuid})
    )
    agent._initial_data_handle = data["handle"]
    agent._initial_data_peek = data

try:
    response = await asyncio.wait_for(
        agent.run(item.query), timeout=AGENT_TIMEOUT_SECONDS
    )
    result.answer = response.content
    result.trajectory = response.trajectory
    result.status = "success"
except asyncio.TimeoutError:
    result.status = "error"
    result.error = "Agent timed out"
\end{lstlisting}

Results are written incrementally---one JSON file per domain---so long runs are
\emph{resumable}: on restart, domains with an existing output file are skipped.
The runner can process multiple capabilities sequentially or in parallel
(\texttt{asyncio.gather}) and optionally streams OpenTelemetry traces to a
Phoenix instance.

\paragraph{Agent Implementation.}
Every model under test is wrapped in a LangGraph ReAct agent
(\texttt{create\_react\_agent}) behind a single abstract \texttt{AgentInterface},
whose \texttt{run} method accepts a query or a multi-turn message list and
returns a structured \texttt{AgentResponse}. Because all models---open and
closed---are driven through this identical interface, measured differences
reflect the models rather than bespoke scaffolding.

\begin{lstlisting}[caption={The uniform agent interface and ReAct construction.}]
class AgentInterface(ABC):
    @abstractmethod
    async def run(self, input: Union[str, List[Message]]) -> AgentResponse:
        """Run the agent; return the final answer plus trajectory."""

class LangGraphReActAgent(AgentInterface):
    def _build_agent(self, tools):
        return create_react_agent(self._llm, tools)
\end{lstlisting}

Models are constructed by a provider-agnostic factory (\texttt{create\_llm})
supporting Anthropic, OpenAI, Ollama, LiteLLM, and watsonx; adding a provider
requires no change to the agent or runner. Tools discovered over MCP are
converted to LangChain \texttt{StructuredTool}s by an \texttt{MCPToolWrapper}
that translates each tool's JSON Schema into a Pydantic model, so the model sees
complete, typed parameter signatures rather than opaque schema blobs. The agent
is never made aware of task characteristics---the number of hops a question
requires, or whether retrieval is needed---and answers given only its tool
collection and a tool-use policy when one is supplied. For large catalogs, an
optional \texttt{ToolShortlister} embeds tool names and descriptions with a
sentence-transformer and retains the top-$k$ most similar tools per query.

\begin{lstlisting}[caption={Optional per-query tool shortlisting.}]
if self._shortlister:
    active_tools = self._shortlister.shortlist(query, self._tools)
    agent = self._build_agent(active_tools)
else:
    active_tools = self._tools
    agent = self._agent
\end{lstlisting}

A configurable iteration limit (mapped to LangGraph's recursion limit) bounds
runaway loops, and a fallback parses tool calls emitted as plain-text JSON for
models without a native tool-calling API. Every run yields a complete
trajectory---reasoning steps, tool calls, arguments, and results---enabling
offline scoring of both final-answer correctness and the tool-use process.

\section{Containerized Execution Environment}
\label{app:environment}

\subsection{Execution Environment and Harness}
\label{app:harness_details}
\paragraph{Self-Hosted Infrastructure.}
All tools ship in a \emph{single Docker image} (\texttt{benchmark\_environ}),
instantiated as one container per capability. The environment hosts
\emph{structured API tools} (SQL queries over domain-specific SQLite databases)
and \emph{retrieval tools} (semantic search over 62 ChromaDB collections
embedded with IBM \texttt{granite-embedding-english-r2}); structured tools are
surfaced as slot-filling/selection interfaces for SEL/SLOT and as REST endpoints
otherwise. Routing uses zero-overhead process replacement: a shared entrypoint
reads \texttt{CAPABILITY\_ID} and \texttt{os.execv()}s into the right MCP server.
REST-backed servers need no hand-written tool definitions---each converts its
FastAPI OpenAPI spec into typed MCP tools and filters to the active domain.
Agents communicate over the Model Context Protocol on stdio; databases and
indices are never exposed, so every response is deterministic, verifiable, and
free of external dependencies.

% \todo{DC: I didn't understand this para}\paragraph{Adversarial Tool Surface.}
% In \name{}, selecting the \emph{right} tool is part of the task: the
% environment deliberately seeds the tool surface with plausible distractors. The
% capability-4 server applies \emph{asymmetric} filtering---structured tools are
% restricted to the primary domain, while retrieval tools also expose confusable
% ``negative'' domains. Combined with domains exposing up to several hundred
% tools, this makes tool grounding a first-class axis of difficulty.

\paragraph{Reproducible, One-Command Setup.}
The environment needs no hosted service, API key, or cloud dependency: the image
and backing data are published to HuggingFace, and \texttt{docker compose up -d}
launches all four containers. To guard against tool-surface drift, we commit a
SHA-256 checksum over tool names and input schemas for each
\texttt{(capability, domain)} pair; both server and runner verify it before any
query, and a mismatch raises a hard error---keeping every reported number
reproducible against a known tool surface.\\

\noindent{\bf Benchmark Runner and Agent Implementation:}
A runner orchestrates the evaluation lifecycle---loading benchmark items,
connecting to each capability container's MCP server via \texttt{stdio\_client},
streaming verified tool definitions, and recording a complete trajectory (tool
calls, responses, final answer) to structured JSON under a fixed per-query
timeout. Every model is wrapped in a LangGraph ReAct agent behind a uniform
\texttt{AgentInterface}, so open and closed models from a provider-agnostic
factory (Anthropic, OpenAI, Ollama, LiteLLM, watsonx) are evaluated identically.
MCP tools are converted to typed LangChain \texttt{StructuredTool}s exposing
complete parameter signatures. The agent is never told task
characteristics---the hops required, or whether retrieval is needed---and
answers given only its tools and tool-use policy, if present
(see Appendix~\ref{app:harness_details}, Table~\ref{tab:main_results}).

\subsection{Docker Environment}
\label{app:docker_env}

The main text describes \name as a self-hosted environment built from a
single Docker image and brought up with one command. This appendix details that
architecture: how the image is built, how data is kept separate from code, how
containers boot, and how MCP servers are spawned on demand.

\paragraph{One image, four containers.}
The entire environment is a \emph{single} image, \texttt{benchmark\_environ},
built once from \texttt{docker/Dockerfile.unified} and instantiated by
\texttt{docker-compose.yml} as one container per capability
(\texttt{capability\_1\_bi\_apis} through \texttt{capability\_4\_multiturn}).
The image bundles all four tool backends---the M3 REST server, the
retriever server, the SEL/SLOT Python-tools server, and the BPO server---so a
capability is selected purely by configuration, never by a different image. A
key consequence: the four task settings are byte-identical in their code and
dependencies, so cross-capability comparisons cannot be confounded by
environment drift.

\paragraph{Code in the image, data mounted at runtime.}
The image carries only server code and dependencies; the SQLite databases and
pre-built ChromaDB indices are \emph{never} baked in. The Dockerfile declares
their locations as volumes, and the data is published separately as a
HuggingFace dataset, downloaded into a local
\texttt{data/} directory and bind-mounted read-only. This keeps the image small,
lets data and code be versioned independently, and makes the build fully
reproducible. One build-time step is worth noting: the Granite embedding model
is downloaded \emph{during the build}, so a running container has no network
dependency and embedding behaviour is pinned to the image.

\begin{lstlisting}[caption={\texttt{docker/Dockerfile.unified} --- pinned embeddings, code-only image, data as volumes.}]
# Pre-download the embedding model so containers need no network at runtime.
RUN python -c "from sentence_transformers import SentenceTransformer; \
    SentenceTransformer('ibm-granite/granite-embedding-english-r2')"

# Data is mounted at runtime, never baked into the image.
VOLUME /app/db
VOLUME /app/retrievers/chroma_data

# Scripts are run as `python /app/.../server.py`; Python only adds the script's
# own directory to sys.path, so set PYTHONPATH so `from environment.*` resolves.
ENV PYTHONPATH=/app
ENTRYPOINT ["/app/entrypoint.sh"]
\end{lstlisting}

\paragraph{Container boot: FastAPI up, then idle.}
On \texttt{docker compose up -d}, each container runs
\texttt{docker/entrypoint-unified.sh}, which starts the internal FastAPI
services and then health-checks them before declaring the container ready. The
M3 REST service (port 8000) is always started; the retriever service (port
8001) starts \emph{only} if a non-empty \texttt{chroma\_data/} directory is
mounted---so the same image runs lean for capabilities 1--3 and fully loaded for
capability~4. The retriever is given a much longer readiness window because it
must load the embedding model into memory. Once services are healthy the
entrypoint blocks on \texttt{tail -f /dev/null}, keeping the container alive as a
host for on-demand \texttt{docker exec} calls.

\begin{lstlisting}[caption={\texttt{docker/entrypoint-unified.sh} --- conditional retriever, health-gated readiness.}]
# M3 REST FastAPI (port 8000) -- always started.
uvicorn app:app --host 0.0.0.0 --port 8000 &

# Retriever FastAPI (port 8001) -- only if ChromaDB data is mounted.
if [ -d "/app/retrievers/chroma_data" ] && [ -n "$(ls -A /app/retrievers/chroma_data)" ]; then
    uvicorn server:app --host 0.0.0.0 --port 8001 &
fi

# Block readiness on a health check; retriever gets a longer timeout
# because it loads the embedding model into memory.
for i in $(seq 1 60);  do curl -sf localhost:8000/openapi.json && break; sleep 1; done
for i in $(seq 1 300); do curl -sf localhost:8001/health     && break; sleep 1; done

# Idle so the container stays alive for `docker exec` MCP spawns.
exec tail -f /dev/null
\end{lstlisting}

\paragraph{MCP servers spawned on demand.}
No MCP server runs persistently. When the benchmark runner needs one, it issues
a \texttt{docker exec} into the relevant container with the
\texttt{CAPABILITY\_ID} (and \texttt{MCP\_DOMAIN}) environment variables, hitting
a single dispatch entrypoint, \texttt{docker/mcp\_dispatch.py}. The dispatcher
reads \texttt{CAPABILITY\_ID} and \texttt{os.execv()}s into the correct server,
replacing its own process so the MCP client talks to the target server directly
with zero proxy overhead. The lifetime of an MCP server is exactly the lifetime
of one domain's evaluation.

\begin{lstlisting}[caption={\texttt{docker/mcp\_dispatch.py} --- one entrypoint, zero-overhead routing.}]
_ROUTES = {
    "1": [sys.executable, "-m", "environment.m3.python_tools.mcp"],
    "2": [sys.executable, "/app/m3-rest/mcp_server.py"],
    "3": [sys.executable, "/app/environment/bpo/mcp/bpo_router.py"],
    "4": [sys.executable, "/app/retrievers/capability_4_mcp_server.py"],
}

cmd = _ROUTES[os.environ["CAPABILITY_ID"]]
os.execv(cmd[0], cmd)   # replace this process -- no proxy in the loop
\end{lstlisting}

\paragraph{One-command lifecycle.}
The \texttt{Makefile} wraps the whole lifecycle. First-time setup is a single
target, \texttt{make setup}, which chains \texttt{download} (fetch data from
HuggingFace) $\rightarrow$ \texttt{build} (build the image) $\rightarrow$
\texttt{test} (smoke test) $\rightarrow$ \texttt{start} (\texttt{docker compose
up -d}) $\rightarrow$ \texttt{validate}. Both the container runtime and the
Python interpreter are auto-detected, so \texttt{docker} and \texttt{podman}
hosts are supported interchangeably. Two verification layers run before any
benchmark: \texttt{make test} spins up a throwaway container and checks that
required files exist and that the MCP servers complete the protocol handshake;
\texttt{make validate} checks live MCP connections against the running
containers. At runtime, the compose file sets \texttt{MCP\_VERIFY\_CHECKSUMS=1},
so the tool-integrity checksums are enforced by default.

\begin{lstlisting}[language=bash,caption={First-time setup --- data, image, containers, verification.}]
make setup     # download -> build -> test -> start -> validate
# equivalently, step by step:
make download  # fetch SQLite + ChromaDB data from HuggingFace into data/
make build     # docker build -t benchmark_environ -f docker/Dockerfile.unified .
make start     # docker compose up -d  (all four capability containers)
python benchmark_runner.py --capability_id 2 --domain hockey
\end{lstlisting}

\noindent
For interactive inspection, an optional compose override
(\texttt{docker-compose.ports.yml}) maps the internal FastAPI ports to the host
so the Swagger UI and OpenAPI specs can be browsed directly; the MCP path used
by the benchmark needs no exposed ports. Together, these pieces realize the
design goal stated in the main text: a self-hosted, deterministic environment
with no external service dependencies that any user can reproduce with a single
command.

%% file: appendix/multiturn_multihop.tex
\section{Multi-Hop Query Generation Pipeline}
\label{app:query_generation}

This section provides full details of the four-stage pipeline summarized in Section~\ref{sec:data-const} (detailed algorithm \ref{alg:dialogue_generation} of the process). The pipeline constructs multi-hop, multi-turn tasks by composing API calls and retrieval steps into reasoning chains of $2$--$5$ hops. Mistral-Large-2411~\cite{mistral_large_2411} is used for all query generation and mistralai/Mixtral-8x22B-v0.1 is used for judging query quality.

\subsection{Knowledge Graph Construction}

Queries from BIRD-SQL~\citep{bird} are parameterized to expose named entities and schema-linked variables. Each entity is mapped to a Wikidata5M~\citep{wang2021kepler} identifier (QID) via string matching and disambiguation against the Wikidata label index. The resulting triples form domain-specific knowledge graphs $G_q = (V, E)$ for each query $q$.

We then derive a \emph{query connectivity graph} by linking queries whose outputs and inputs share compatible entities: if query $q_a$ returns an entity $e$ that appears as a required parameter of query $q_b$, we add a directed edge $(q_a, q_b)$. This graph encodes all feasible multi-hop API chains within and across domains.

\subsection{Dialogue Trajectory Generation}

Multi-turn dialogues are generated via depth-first traversal over the query connectivity graph, motivated by the compositional question generation approach of \citet{trivedi2022musique}. At each traversal step, a hop count $h \in \{1, 2, 3\}$ is sampled with weights $(0.10, 0.60, 0.30)$, biasing the dataset toward two-hop reasoning chains while still including shorter and longer chains for diversity. Traversal segments are sequentially composed into trajectories of up to $7$ turns representing chained API reasoning.

For each candidate trajectory, an LLM generates a natural-language question that requires executing the full chain to answer. The generated question is evaluated by an LLM judge for:
\begin{itemize}
    \item \textbf{Answerability}: The question must require all hops in the chain (no shortcuts).
    \item \textbf{Naturalness}: The question must read as a plausible user query, not a mechanical composition of sub-queries.
    \item \textbf{Groundedness}: The answer must be derivable from the API responses in the chain.
\end{itemize}
Questions failing validation are rewritten (up to $2$ attempts) or discarded.

\subsection{Retrieval Augmentation}

To produce API+RAG tasks, we introduce retrieval-based edges into the connectivity graph. For knowledge graph relations connecting entities $e_i$ and $e_j$, the system retrieves Wikipedia passages (from ClapNQ~\citep{rosenthal2025clapnq}) associated with the corresponding Wikidata QIDs and extracts paragraphs mentioning both linked entities. An LLM then generates a combined question that requires both an API call and a retrieval step, validated for groundedness and naturalness as above.

This produces reasoning chains with mixed source-type patterns such as API$\to$RAG, RAG$\to$API, and API$\to$RAG$\to$API, where retrieval outputs directly condition subsequent API parameters or vice versa.

\subsection{Retrieval-Only Multi-Hop Dialogues}

To complement the API+RAG setting, we generate retrieval-only multi-turn dialogues following the methodology of \citet{lee2024multidocumentgroundedmultiturnsynthetic}. A \emph{user} agent generates entity-grounded questions from domain-specific ClapNQ documents, interleaved with \emph{agent} responses. Each query is validated against the domain database schema to ensure it cannot be answered via structured APIs---API-answerable queries are removed as part of cross-source answerability filtering. Responses are additionally verified for groundedness against retrieved passages.

To introduce multi-hop structure, sequential question-answer pairs are merged by an LLM into a single bridging query that omits the intermediate answer, producing patterns of the form \texttt{(RAG-RAG)(RAG)$\cdots$} or \texttt{(RAG)(RAG-RAG)$\cdots$}. The resulting dialogues form clean retrieval-only trajectories ensuring cross-source unanswerability. Full details of this sub-pipeline appear in Appendix~\ref{app:multi_rag_pipeline}.

\subsection{Retrieval Index Construction}
\label{appsubsection:retrieval_index_construction}

Domain-specific retrieval indices are constructed from documents sourced from Wikidata5M and ClapNQ. To minimize cross-source answerability (i.e., ensuring RAG questions cannot be answered via APIs and vice versa), we apply an LLM-based filtering procedure:
\begin{enumerate}
    \item Documents capable of answering any existing API query are identified and removed from the retrieval corpus.
    \item RAG queries answerable via structured APIs are discarded from the task set.
    \item Cross-domain contamination is checked: documents that can answer queries from unrelated domains are removed.
    \item Surviving documents are sampled to $20{,}000$ per domain and indexed using ChromaDB, forming the retrieval tools for document-grounded tasks.
\end{enumerate}

\begin{algorithm}
\caption{\name\ Dialogue Generation}
\label{alg:dialogue_generation}
\begin{algorithmic}[1]
\REQUIRE BIRD query set $Q$ from domain $D$, Wikidata5M triples $W$, document corpus $C$
\ENSURE Multi-hop multi-turn dialogue dataset $M$

\STATE $M \leftarrow \emptyset$

\FOR{each query $q \in Q$}
    \STATE Extract named entities $E_q$
    \STATE Construct domain knowledge graph\\ $G_q=(V,E)$ using triples from $W$
    \STATE Build query connectivity graph using answer--parameter links and RAG links
    \STATE Initialize dialogue trajectory $T \leftarrow \emptyset$

    \STATE Generate candidate traversal paths by DFS

    \FOR{each traversal step $i$}
        \STATE Sample hop count $h_i \sim \{1,2,3\}$ with weights $(10,60,30)$
        \STATE Generate traversal segment $P_i$
        \IF{$P_i$ corresponds to a RAG relation}
            \STATE Retrieve Wikipedia passages using entity QIDs
            \STATE Generate API-RAG question
        \ELSE
            \STATE Generate API multi-hop question
        \ENDIF

        \STATE Evaluate question with LLM-Judge
        \IF{question is valid}
            \STATE Append question to trajectory $T$
        \ELSE
            \STATE Rewrite or discard question
        \ENDIF
    \ENDFOR

    \STATE Segment trajectory $T$ into dialogues of at most 7 turns
    \STATE Add resulting dialogues to dataset $M$

\ENDFOR

\RETURN $M$
\end{algorithmic}
\end{algorithm}

% \subsection{Query Generation Prompts \& Judges}

%% file: appendix/multiturn_rag.tex
\section{Multi-Turn RAG Benchmark Data Generation Pipeline}
\label{app:multi_rag_pipeline}

This section describes our automated pipeline for generating high-quality multi-turn conversational RAG (Retrieval-Augmented Generation) benchmark data. The pipeline employs multiple AI agents with sophisticated quality control mechanisms to ensure data integrity and prevent contamination. Following automated dialogue and query generation, four human annotators iteratively filter out samples that do not satisfy the desired evaluation rigor, including issues related to ambiguity, insufficient grounding, or limited reasoning complexity. The final benchmark consists of 100 carefully curated dialogues retained after this annotation and validation process.

\subsection{Overview}

The Multi-RAG Pipeline is an automated system that generates multi-turn conversational data for evaluating RAG systems across multiple domains. The pipeline orchestrates interactions between specialized agents while applying rigorous quality checks at each step to ensure the generated data is clean, grounded, and suitable for benchmarking purposes.

\subsection{System Architecture}

The pipeline consists of four primary components:

\begin{itemize}
    \item \textbf{User Agent}: Simulates realistic user queries with diverse question types
    \item \textbf{RAG Agent}: Generates responses based on retrieved documents
    \item \textbf{Answer Selection Agent}: Validates and selects appropriate responses
\end{itemize}

\subsubsection{Agent Configuration}

Each agent is configured with specific parameters:

\begin{itemize}
    \item \textbf{Model Provider}: Supports OpenAI models
    \item \textbf{Temperature}: Set to 0.0 for deterministic RAG responses and LLM judege, 1.0 for diverse user queries
    \item \textbf{Max Tokens}: 4096 tokens for comprehensive responses
    \item \textbf{Sampling}: User agent uses $n=3$ for generating multiple query candidates (see Appendix~\ref{app:entity_answer_selection}) 
\end{itemize}

\subsection{Data Generation Process}

The pipeline generates multi-turn conversations through the following iterative process:

\subsubsection{Conversation Initialization}

For each conversation sample:
\begin{enumerate}   
    \item Set Elasticsearch index to \texttt{ClapNQ}, configure domain name, description and keywords for domain-specific context retrieval.
    \item Initialize conversation with domain-specific context from \texttt{ClapNQ}
    \item Reset conversation history and metadata
\end{enumerate}

\subsubsection{Turn Generation Loop}

Each turn in the conversation follows this workflow:

\paragraph{Step 1: User Query Generation}
The User Agent generates a query based on:
\begin{enumerate}
    \item Retrieved documents from the domain-specific corpus plus Conversation history from previous turns
    \item Determine query type (entity-based, factoid, etc.)
\end{enumerate}

The agent employs sophisticated prompts (see Appendix~\ref{app:user_prompts}) and conversation, document context to generate questions including:
\begin{itemize}
    \item \textbf{Entity questions}: Queries whose answer is a specific named entity.(see Appendix~\ref{app:entity_question})
    \item \textbf{Factoid questions}: Seeking brief, factual information for the name entity mentioned above.
\end{itemize}

\paragraph{Step 2: RAG Response Generation}
The RAG Agent:
\begin{enumerate}
    \item Receives the user query concatenated with retrieved passages.
    \item Generates response to the user query. Uses the prompt template shown in Appendix~\ref{app:rag_prompts}
\end{enumerate}

\paragraph{Step 3: Multi-hop Query Merging}
For creating multi-hop queries, the system can merge sequential question-answer pairs:
\begin{enumerate}
    \item Randomly selects merge position (turn 1 or turn 3)
    \item Extracts first query, answer, and second query from selected turns
    \item Uses LLM to generate merged query that combines both questions without mentioning intermediate answer
    \item Validates naturalness of merged query
    \item Creates new conversation with merged query replacing original turns with multi-hop patterns: \texttt{(RAG-RAG)(RAG)(RAG)} or \texttt{(RAG)(RAG)(RAG-RAG)}
\end{enumerate}

\paragraph{Step 4: Data Recording}
Upon successful validation, the system records:
\begin{itemize}
    \item User query and metadata (entity, document, query type)
    \item RAG response and supporting documents
    \item Gold sequence with retriever information and retrieved document chunks
\end{itemize}

\subsection{Quality Control Mechanisms}

The pipeline implements multiple layers of quality control to ensure data integrity and prevent contamination.

\subsubsection{In-Generation Quality Checks}

Quality checks applied during the generation process (between Steps 1-3):

\begin{enumerate}
    \item \textbf{Answerability Check}:
    \begin{itemize}
        \item Constructs conversation history including the new query
        \item Uses LLM judge (see Appendix~\ref{app:api_check}) to analyze whether RAG queries can be answered by querying structured databases
        \item Uses LLM judge (see Appendix~\ref{app:doc_check}) to analyze whether API queries can be answered by retriving documents. 
        \item Rejects/drops the conversation if the RAG query can be answered via SQL/API calls, also vice versa. 
        \item Ensures pure RAG evaluation by preventing API-answerable contamination, and vice versa. 
    \end{itemize}

    \item \textbf{Groundedness Verification}: Verify answers to the queries are factually supported by retrieved documents using LLM-based validation
    
    \item \textbf{Unanswerability Detection}: Rejects responses containing "I can not answer" or reject answering the question.

    \item \textbf{Conversation Rejection}: Discards entire conversation if any quality check fails
    
    \item \textbf{Completeness Check}: Ensures conversations reach the target number of turns (default: 6)
\end{enumerate}

\subsubsection{Post-Generation Decontamination}

After initial data generation, comprehensive decontamination ensures no data leakage:

\begin{enumerate}
    \item \textbf{Cross-Domain Filtering}:
    \begin{itemize}
        \item Identifies documents that can answer questions from other domains
        \item Removes samples with contaminated documents
    \end{itemize}
    
    \item \textbf{ClapNQ Decontamination} (per domain):
    \begin{itemize}
        \item Retriever corpus for each domain is combinations of ClapNQ and ground truth document chunks for RAG question, this ensures no document appears in both sets.
        \item Outputs final cleaned ClapNQ documents for each domain
    \end{itemize}
    
    \item \textbf{Domain-ClapNQ Sampling}:
    \begin{itemize}
        \item Randomly samples cleaned ClapNQ documents to reach 20,000 total per domain.
        \item Saves merged 20k document collections per domain as final retrieval corpus.
    \end{itemize}
\end{enumerate}

\subsection{Output Format}

The pipeline generates data in M3 benchmark format with the following structure:

\begin{lstlisting}
{
  "task_name": "domain_name",
  "dataset_name": "domain_name",
  "sample_id": 0,
  "turns": [
    {
      "query": "user question",
      "answer": "rag response",
      "type": "(RAG)",
      "gold_sequence": [
        {
          "question": "user question",
          "answer": "rag response",
          "rag_doc": ["doc1", "doc2", ...],
          "question_type": "RAG",
          "db_id": "domain",
          "output": [
            {
              "name": "retriever_clapnq_domain",
              "arguments": {"query": "user question"}
            }
          ],
          "OUTPUT_AFTER_EXECUTING_API": ["doc1", "doc2"]
        }
      ],
      "metadata": {
        "query": {...},
        "answer": {...},
        "user_query_type": "entity_as",
        "entity": "EntityName"
      }
    }
  ],
  "num_turns": 6,
  "num_hops": [1, 1, 1, 1, 1, 1],
  "type": "(RAG)(RAG)(RAG)(RAG)(RAG)(RAG)"
}
\end{lstlisting}

\subsection{Domain Coverage}

The pipeline supports 47 domains including:
\begin{itemize}
    \item \textbf{Entertainment}: disney, movies, shakespeare, simpson-episodes
    \item \textbf{Sports}: olympics, professional-basketball, european-football, hockey
    \item \textbf{Geography}: mondial-geo, world, address
    \item \textbf{General Knowledge}: books, video-games, restaurant, chicago-crime
\end{itemize}

\subsection{Prompts}
\label{app:user_prompts}

\subsubsection{First Turn - Entity Question Prompt}
\label{app:entity_question}
{\footnotesize
\begin{lstlisting}
You will be given a document. And you need
generate a question and its answer is
exactly the one proper noun entity from
this document. And the answer can not be
other entities mentioned in the documents.

Below are instructions:
1. Identify one name entity from the
   document, and this entity should not be
   among a serial of entities (people's
   name) in the documents.

2. Create one question whose correct
   answer is exactly this named entity.

3. The answer must be the entity name
   only, with no extra words,
   explanations, or punctuation. And can
   not be other entities.

4. The question must be answerable only
   from the document.

You should generate following below format:
<question>
Generate the question and its answer is
exactly the one proper noun entity. And
the answer can not be other entities
mentioned in the documents.
The question should be concise enough and
do not require clarification of "This",
"That", "These"... nor other co-reference
words.
And Make sure there is no "one of"
appeared in the question.
If you can not generate such question,
write "I can not generate"
</question>

<entity>
Write the entity of the answer to the
question you generated.
</entity>
\end{verbatim}
}

\subsubsection{Follow-up Turn - Entity Fact Question Prompt}
\label{app:entity_fact_question}
{\footnotesize
\begin{verbatim}
You are a helpful assistant.
Given context of document starting with
<document> and end with </document>
following by conversation history of user
query and assistant responses, your task
is to generate a concise follow-up
question.

To generate a subsequent user query, think
step-by-step:
1. Read the document and understand its
   content.
2. Identify the main topic or the key
   point being discussed in the
   conversation.
3. Read understand the given user queries
   and an agent responses in the
   conversations.
4. Based on provided documents and
   conversation history, you should
   generate a question about the proper
   noun name entity appeared in the
   document.

Please generate output following below
format:

<question>
generate question considering below rules:
1. The question is asking some fact of
   the entity itself, instead of its
   belongings nor other proper noun
   entity.
2. The question should not start with
   "what other", please ask more
   straightforward question without double
   checking previous context for its
   answer.
3. The question should be concise enough
   and do not require clarification of
   "This", "That", "These"... nor other
   co-reference words.

If you can not generate such question,
write "I can not generate"
</question>
\end{lstlisting}
}

\subsubsection{Follow-up Turn - Entity Answer Selection Prompt}
\label{app:entity_answer_selection}
{\footnotesize
\begin{lstlisting}
You are a helpful assistant.
Given context of document starting with
<document> and end with </document>
following by conversation history, your
task is to generate a concise follow-up
question.

To generate a subsequent user query, think
step-by-step:
1. Read the document and understand its
   content.
2. Identify the main topic or the key
   point being discussed in the
   conversation.
3. Read understand the given user queries
   and responses.
4. Based on provided documents and
   conversation history, you should do
   below two tasks:

task1: try to find one proper noun name
entity (names start with capital letter)
appear in the documents, but it did not
appear in the conversation history. And
this entity should not be among a serial
of entities (people's name) in the
documents.

task2: if task1 succeed, choose one such
entity, and generate a question and the
answer of this question is exactly this
entity, not others.

The generated question should be related
to the main topic of the current
conversation and naturally follow up the
last turn of the conversation.
Never ask question mentioned "one of" in
it.

Please generate output following below
format:

<question>
Generate the question and its answer is
exactly entity appeared in the document.
And this entity is the only answer to this
question.
The question should be concise enough and
do not require clarification of "This",
"That", "These"... nor other co-reference
words.
If you can not generate such question,
write "I can not generate"
</question>

<entity>
the entity you used to generate above
question as the answer to this question.
</entity>
\end{lstlisting}
}

\subsubsection{RAG Agent Prompt}
\label{app:rag_prompts}
{\footnotesize
\begin{lstlisting}
System: You are helpful assistant who
answers users' queries based on relevant
documents provided

User: Please read below documents:
<<documents>>

Please answer this question:
<<query>>

After you understand above documents and
question, You should wrote your answer
following below format:

<explanation>
you should explain step by step how you
should look for the answer in the
documents. The explanation should repeat
the question, and reason how you should
look for the answer in the documents. Also
reason why certain pieces of information
in the document is relevant to answer the
question, and whether the information in
multiple sections of the document can be
combined to identify the answer.
</explanation>

<detailed_answer>
generate a detailed long answer to the
question based on provided documents. If
the answer to the query is not available,
you should generate "I am sorry, the
question is unanswerable from the
available information" with corresponding
reasoning.
The answer must be factually coherent to
the document and must not include any
facts that are not found in the document.
The answer should contain a summary of
reasoning. Imagine that the user is not
aware that you are reading a document to
answer the question. Hence do not mention
the word "document" in the answer.
</detailed_answer>

<final_answer>
Generate a one or two concise sentence
here of final answer by summarizing the
detailed answer. The summary must contain
the most critical points of the detailed
answer. Imagine that the user is not aware
that you are reading a document to answer
the question. Hence do not mention the
word "document" in the answer.
If you can not directly answer the
question based on the document, generate
"I can not answer." here.
</final_answer>

<consistency>
reason whether the answer and the
explanation are consistent. Generate the
reasoning for the consistency evaluation
and then provide a yes or no answer for
consistency.
</consistency>

Please follow above instruction and
generate your response.
\end{lstlisting}
}

\subsubsection{API Answerability Check Prompt}
\label{app:api_check}
{\footnotesize
\begin{lstlisting}
You are given a database schema in CSV
format describing several tables. Each
schema entry includes 5 columns:

"original_column_name", "column_name",
"column_description", "data_format",
"value_description".

You are also given a conversation history
between a human and an SQL assistant, the
last turn of the conversation is the
user's query, your task is to determine
whether this query can be possibly
answered by writing SQL to query the
database.

Below is the database schema start with
<schema> and end with </schema>:
<schema>
{schema}
</schema>

And below is the conversation history
start with <conv> and end with </conv>:
<conv>
{conv}
</conv>

Now please analyze the schema and
conversation context above, assuming the
previous user query has been answered
properly by the assistant (no need to
verify them using SQL any more).
and you should determine if the last turn
of user query is answerable or not, using
the conversation history as context.

Please generate following below format:

<reasoning>
Write an explanation of your decision
based on the schema
</reasoning>

<canAnswer>
Whether the query can be possibly answered
by writing SQL based on schema or not.
Write "yes" if it can, otherwise write
"no".
</canAnswer>

<SQL>
Write the SQL query if canAnswer is true;
if the table name contains hyphen (-) use
backticks (`) to enclose the table name,
otherwise write "None"
You should use original column names when
writing SQL.
You could apply basic operations on
provided tables and columns: Querying,
Filtering, Sorting, Joins and Aggregation.
You should Derive table names from the CSV
filenames (e.g., director.csv -> director)
</SQL>
\end{lstlisting}
}

\subsubsection{Document Answerability Check Prompt}
\label{app:doc_check}
{\footnotesize
\begin{lstlisting}
The following tasks each contains a
document, a conversation and a response to
the last turn of the conversation.
The response is supposed to rely on the
document for its source of information,
optionally using common sense knowledge
and common sense inference, but it may
fail this, and instead contain substantial
claims that are not grounded in the
document or common sense knowledge.

Your task is to assess whether the
response is entirely grounded in the
document, grounded in the document plus
common sense knowledge and reasoning, or
ungrounded. To make this determination,
perform the following steps:
1. Identify all substantial claims in the
   response:
   - Ignore non-substantial claims, such
     as greetings or self-descriptions
     such as "I'm a helpful assistant",
   - Try to formulate each claim in a
     stand-alone form with all pronouns
     and other references resolved;
2. Assess the grounding of each of these
   claims:
   - If it is essentially a rephrasing of
     information from the document, or can
     be derived from such information by
     trivial common-sense reasoning, it is
     grounded, This is so even if it
     contradicts other parts of the
     document.
   - If it relies on, in additional to
     information from the document,
     additional non-trivial common sense
     knowledge or common sense reasoning,
     it is partially grounded,
   - If a claim is about the provided
     document, or about the agent's state
     of knowledge, with the effect of not
     being able to answer the user
     inquiry, it is grounded if and only
     if the required information is indeed
     lacking in the document.
   - If a claim cannot be derived directly
     from the document or indirectly with
     help of common sense knowledge and
     reasoning, it is ungrounded;
3. Make the overall decision according to:
   - If at least one claim is not
     grounded, the response is not
     grounded (Note that this is not a
     case of partially grounded);
   - Otherwise if at least one claim is
     partially grounded, the response is
     partially grounded;
   - Otherwise the response is grounded.

Pay attention that: Even if the document
contains the keyword of response, it does
not mean the response is grounded, and you
have to make decision based on 1,2,3
above.

Your final conclusion should be written in
two lines:
- The first line contains one of the
  following labels [yes, partial, no,
  unsure],
  - "yes" is for grounded,
  - "partial" is for grounded with
    non-trivial common sense knowledge or
    reasoning,
  - "no" is for ungrounded,
  - "unsure" is for the situations where
    the document, conversation or response
    contain ambiguities such that
    different interpretations lead to
    different conclusions about
    groundedness;

Here is the document starting with <doc>
and end with </doc>
<doc>
{doc}
</doc>

Here is the conversation starting with
<conv> and end with </conv>.
<conv>
{conv}
</conv>

Here is the response to the last turn of
conversation starting with <response> and
end with </response>.
<response>
{response}
</response>

Now write your final conclusion following
below format:
<conclusion>
choose a label from [yes, partial, no,
unsure] based on your analysis of given
document and response.
- "yes" is for grounded,
- "partial" is for grounded with
  non-trivial common sense knowledge or
  reasoning,
- "no" is for ungrounded,
- "unsure" is for the situations where the
  document, conversation or response
  contain ambiguities such that different
  interpretations lead to different
  conclusions about groundedness;
</conclusion>
\end{lstlisting}
}

\begin{algorithm}[H]
\caption{Dialog Generation for Document Retrieval Tasks}
\label{alg:retrieval_enhancement}
\begin{algorithmic}[1]
\REQUIRE Domain set $\mathcal{D}$, ClapNQ index $\mathcal{E}$, conversation length $L$
\ENSURE Multi-turn RAG dialogue dataset $\mathcal{R}$
\STATE $\mathcal{R} \leftarrow \emptyset$
\FOR{each domain $d \in \mathcal{D}$}
    \STATE Configure $\mathcal{E}$ with domain name, description, and keywords for $d$
    \STATE Initialize conversation history $H \leftarrow \emptyset$, metadata $\leftarrow \emptyset$
    \FOR{turn $t = 1$ to $L$}
        \STATE \textit{// Step 1: User Query Generation}
        \STATE Retrieve passages $P_t$ from $\mathcal{E}$ conditioned on $H$
        \STATE Determine query type $\tau_t \in \{\text{entity}, \text{factoid}\}$
        \STATE Generate $n{=}3$ candidate queries $\{q_t^1, q_t^2, q_t^3\}$ using user agent prompted with $P_t$ and $H$
        \STATE Select final query $q_t$ via entity/answer selection (Appendix~\ref{app:entity_answer_selection})
        \STATE \textit{// Step 2: RAG Response Generation}
        \STATE Generate response $r_t$ conditioned on $q_t \oplus P_t$
        \STATE \textit{// Step 3: Record}
        \STATE Append $(q_t, r_t, P_t, \tau_t)$ to $H$
    \ENDFOR
    \STATE \textit{// Step 4: Multi-hop Merging (optional)}
    \STATE Sample merge position $m \sim \{1, 3\}$
    \IF{merge selected}
        \STATE Extract $(q_m, r_m, q_{m+1})$ from $H$
        \STATE Generate merged query $q^* \leftarrow \textsc{Merge}(q_m, r_m, q_{m+1})$ omitting $r_m$
        \IF{$q^*$ passes naturalness validation}
            \STATE Replace turns $m, m{+}1$ in $H$ with multi-hop pattern \texttt{(RAG-RAG)(RAG)} or \texttt{(RAG)(RAG)(RAG-RAG)}
        \ENDIF
    \ENDIF
    \STATE $\mathcal{R} \leftarrow \mathcal{R} \cup \{H\}$
\ENDFOR
\RETURN $\mathcal{R}$
\end{algorithmic}
\end{algorithm} 
\label{sec:appendix}